\documentclass[10pt,journal,compsoc]{IEEEtran}
\usepackage[nocompress]{cite}
\usepackage{ragged2e}
\usepackage{hyperref}
\hypersetup{colorlinks,linkcolor={blue},citecolor={blue},urlcolor={red}}
\usepackage{amsmath,amsfonts}
\usepackage{algorithmic}
\usepackage{algorithm}
\usepackage{array}
\usepackage[caption=false,font=normalsize,labelfont=sf,textfont=sf]{subfig}
\usepackage{textcomp}
\usepackage{stfloats}
\usepackage{url}
\usepackage{verbatim}
\usepackage{graphicx}
\usepackage{graphics}
\usepackage{cite}
\usepackage{ragged2e}
\usepackage[table]{xcolor}
\usepackage{colortbl}
\usepackage{hyperref}
\usepackage{booktabs}
\usepackage{multirow}
\usepackage{colortbl}
\definecolor{tablecl}{gray}{0.9}
\usepackage{color,xcolor}

\def\eg{\emph{e.g.}}
\def\etc{\emph{etc}}
\def\etal{{\em et al.~}}

\begin{document}

\title{Towards Real Zero-Shot Camouflaged Object Segmentation without Camouflaged Annotations}

\author{ Cheng Lei, 
    Jie Fan, 
    Xinran Li, 
    Tian-zhu Xiang, 
    Ao Li,  
    Ce Zhu,
    Le Zhang
    
\thanks{C. Lei, J. Fan, X. Li, A. Li, C. Zhu and L. Zhang are with University of Electronic Science and Technology
of China.}
\thanks{T.-Z. Xiang is with the Space42, UAE.}
\thanks{C. Lei and J. Fan contribute equally.}
\thanks{L. Zhang and C. Zhu are the corresponding authors.}
\thanks{This work was supported by  the Key Program for International Cooperation of Ministry of Science and Technology of China (No.2024YFE0100700) and the Key Project of the Natural Science Foundation of Sichuan Province (No.2025ZNSFSC0002).}

}



\IEEEtitleabstractindextext{
\begin{abstract}
\justifying
Camouflaged Object Segmentation (COS) faces significant challenges due to the scarcity of annotated data, where meticulous pixel-level annotation is both labor-intensive and costly, primarily due to the intricate object-background boundaries. Addressing the core question, "Can COS be effectively achieved in a zero-shot manner without manual annotations for any camouflaged object?", we propose an affirmative solution. We examine the learned attention patterns for camouflaged objects and introduce a robust zero-shot COS framework. Our findings reveal that while transformer models for salient object segmentation (SOS) prioritize global features in their attention mechanisms, camouflaged object segmentation exhibits both global and local attention biases.
Based on these findings, we design a framework that adapts with the inherent local pattern bias of COS while incorporating global attention patterns and a broad semantic feature space derived from SOS. This enables efficient zero-shot transfer for COS. Specifically, We incorporate a Masked Image Modeling (MIM) based image encoder optimized for Parameter-Efficient Fine-Tuning (PEFT), a Multimodal Large Language Model (M-LLM), and a Multi-scale Fine-grained Alignment (MFA) mechanism. 
The MIM encoder captures essential local features, while the PEFT module learns global and semantic representations from SOS datasets. To further enhance semantic granularity, we leverage the M-LLM to generate caption embeddings conditioned on visual cues, which are meticulously aligned with multi-scale visual features via MFA. This alignment enables precise interpretation of complex semantic contexts. Moreover, we introduce a learnable codebook to represent the M-LLM during inference, significantly reducing computational demands while maintaining performance. Our framework demonstrates its versatility and efficacy through rigorous experimentation, achieving state-of-the-art performance in zero-shot COS with $F_{\beta}^w$ scores of 72.9\% on CAMO and 71.7\% on COD10K. By removing the M-LLM during inference, we achieve an inference speed comparable to that of traditional end-to-end models, reaching 18.1 FPS. 
Additionally, our method excels in polyp segmentation, and underwater scene segmentation, outperforming challenging baselines in both zero-shot and supervised settings, thereby implying its potentiality in various segmentation tasks. The source code will be made available at \url{https://github.com/AVC2-UESTC/ZSCOS-CaMF}.
\end{abstract}

}

\maketitle

\IEEEpeerreviewmaketitle


\section{Introduction}
\label{sec:intro}

\begin{figure*}[t]
\centering
	\includegraphics[width=0.99\textwidth]{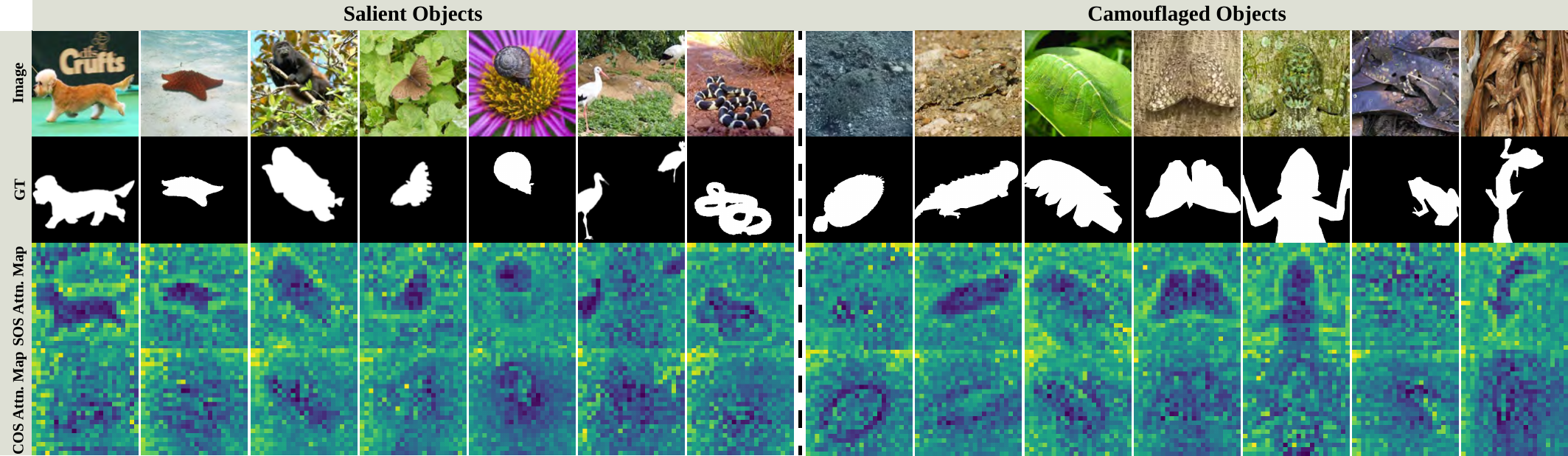}
	\caption{{
		\textbf{Examples of salient and camouflaged objects.} In salient object segmentation (SOS), objects are easily distinguishable from the background, while in camouflaged object segmentation (COS), objects blend into their surroundings. {The attention maps reveal that the SOS model exhibit a global bias, activating broadly over salient objects and most camouflaged objects, with only minor failures. In contrast, the COS model primarily emphasizes local patterns on camouflaged objects while also responding to some salient objects, apart from a few detection inaccuracies. }
		}}
	\label{fig:obv}
\end{figure*}

Camouflaged Object Segmentation (COS) is an emerging computer vision task focused on accurately segmenting objects that seamlessly blend into their surrounding environments \cite{SINet, SINetV2}. Unlike existing segmentation tasks such as salient object segmentation (SOS), where segmented objects are distinctly different from the background and each other, as illustrated in the rows of Figure~\ref{fig:obv}, COS presents a significant challenge due to the inherent visual similarity between foreground objects and the background. This similarity makes it difficult to identify distinguishing cues for accurate separation. Existing COS methods often rely heavily on large-scale annotated datasets. However, creating and accessing such datasets pose significant challenges, primarily due to the demanding nature of extensive pixel-level annotations, which are both time-consuming and labor-intensive~\cite{NC4K, CRNet}.

\begin{figure*}[th]
	\centering
	\includegraphics[width=0.99\linewidth]{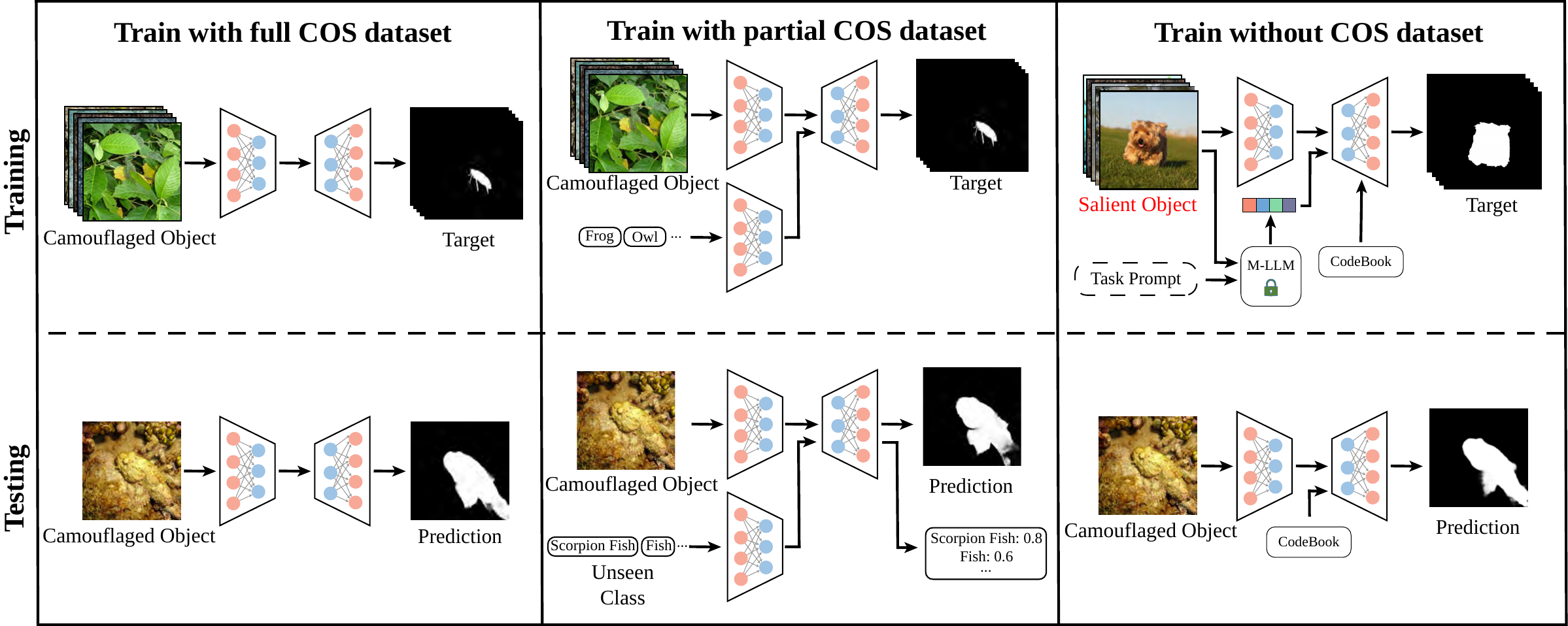}

\begin{minipage}[h]{1.0\linewidth}
\hspace{18pt}
\textbf{(a) Supervised}~\cite{ZoomNet,SINet,SINetV2,FSPNet} \hspace{6pt}
\textbf{(b) Existing ZS} \& \textbf{OV COS}~\cite{ZSCOD,OVCOS} \hspace{34pt}
\textbf{(c) Our ZS Setting}
    \end{minipage} 
 
	\caption{{\bf Overview of different learning pipeline to COS.} In Supervised COS (a), camouflaged objects are used for both training and testing~\cite{SINet, SINetV2, ZoomNet,FSPNet}. In Zero-Shot COD~\cite{ZSCOD} \& Open-Vocabulary~\cite{OVCOS} (b), training data consists of a specific set of camouflaged objects, while testing data includes unseen camouflaged objects. Our Zero-shot COS (c) setting offers a more practical approach as it is trained without any camouflaged annotations. Instead, we utilize a saliency dataset, which is more readily available, cost-effective.}
	\label{fig:overview}

\end{figure*}

Efforts to mitigate these challenges have led to the development of weakly-supervised methods utilizing scribbles or points, aiming to reduce annotation time \cite{CRNet, SAM-MFG}. However, these methods remain time-consuming, especially when dealing with large-scale datasets. In such contexts, zero-shot learning emerges as a crucial approach, eliminating the need for explicit data labeling for downstream tasks.
Existing methodologies such as zero-shot camouflaged object detection (ZSCOD)~\cite{ZSCOD} and open-vocabulary camouflaged object segmentation (OVCOS)~\cite{OVCOS} focus on identifying specific classes of objects, aiming to segment unseen classes as depicted in Figure~\ref{fig:overview}. However, both ZSCOD and OVCOS still require partial data labeling from the COS dataset for training, which does not fully eliminate the need for explicit annotations \cite{ZSCOD, OVCOS}. This limitation highlights the importance of developing a more pragmatic, real-world approach to zero-shot camouflaged object segmentation (ZSCOS) that minimizes reliance on the COS dataset.

Recent developments by GenSAM~\cite{gensam} and MMCPF~\cite{MMCPF} have made strides in addressing these issues by employing prompt engineering to synergize SAM~\cite{SAM} with M-LLM for ZSCOS. These methods effectively utilize the capabilities of pre-trained models and circumvent the necessity for extensive post-training, thereby streamlining the deployment process. 
{However, as shown in Table~\ref{tab:gensam_mmcpf_camf}, they continue to encounter substantial computational overhead during inference, which may limit their practical applicability in resource-constrained settings. This underscores the need for further exploration into solutions that are both efficient and effective, aiming to fully address the constraints of existing methodologies.}

\begin{figure*}[t]
\centering
	\includegraphics[width=0.9\textwidth]{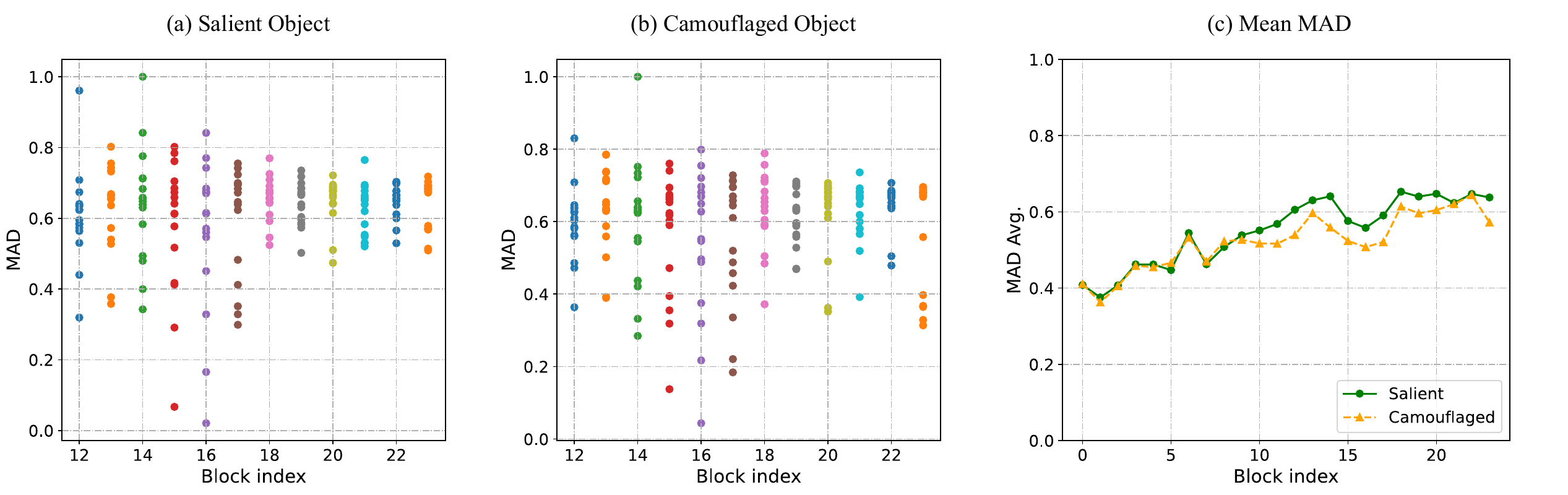}
	\caption{{
		\textbf{Comparison of normalized mean attention distance (MAD) between salient and camouflaged objects. } 
        {(a) and (b) show MAD distributions across attention heads (each point represents one head) for salient and camouflaged objects, respectively. Higher MAD values indicate larger receptive fields, while greater point separation reflects higher diversity among heads. 
        COS model employs both local and global patterns, evidenced by a balanced MAD distribution beyond the 15th layer (some heads with MAD $<$ 0.5, others with MAD $>$ 0.5). In contrast, model trained on salient objects relies primarily on global patterns, with MAD values mostly greater than 0.5. (c) shows that after the 10th layer, camouflaged objects exhibit lower mean MAD values than salient objects.}
		}}
	\label{fig:motiv}
\end{figure*}

Recall that our zero-shot COS task is inherently a zero-shot dense prediction problem. In zero-shot learning, recent studies have shown that the ability to extract robust semantic features is crucial for success~\cite{CLIP}. On the other hand, for dense prediction tasks, effectively processing high-frequency details is essential to accurately capture fine object boundaries~\cite{CLandMIM}. This dual challenge has led us to hypothesize that combining a training dataset rich in semantic information with a pre-trained model proficient in dense prediction could significantly benefit our zero-shot COS approach. By doing so, the system not only improves its ability to capture high-frequency details, essential for object boundary delineation, but also enhances its understanding of semantic features, which are critical for zero-shot learning in dense prediction contexts.
\begin{table}[t]
\centering
\caption{A comparison of inference time and model parameters across GenSAM, MMCPF, and CAMF (our method) on RTX 4060Ti.}
\setlength{\tabcolsep}{2.5pt}
\resizebox{0.99\columnwidth}{!}{
    \begin{tabular}{@{}c|c|c|c@{}}
\toprule
Method      & Inference Macro Structure & Total Parameters     & Inference Time (s/img.)                  \\ \midrule
GenSAM~\cite{gensam}      & M-LLM + VLM + Segmentor   & 7.8 B + 86 M + 636 M & (0.570 + 0.011)$\times$5 + 0.774 = 3.679 \\
MMCPF~\cite{MMCPF}       & M-LLM + Segmentor         & 7 B + 636 M          & 2.648 + 0.797 = 3.445                    \\
CAMF (Ours) & Segmentor + Codebook      & 310 M + 0.8 M        & 0.055                                    \\ \bottomrule
\end{tabular}
}
\label{tab:gensam_mmcpf_camf}
\vspace{-0.2cm}
\end{table}\\

An empirical observations, illustrated in Figure~\ref{fig:obv}, offer additional insights for COS. Specifically, we extract the attention map from the final transformer block of the image encoder, which provides a two-dimensional visualization of the attention weights. These weights are computed by the dot product between the query matrix and the corresponding value matrix, and subsequently averaged across all attention heads. The analysis reveals that while SOS models primarily focus on low-frequency features within semantically significant areas, COS models excel in identifying high-frequency features critical for outlining the edges of camouflaged objects.

Further analysis in Figure~\ref{fig:motiv} presents attention patterns between salient and camouflaged objects during supervised fine-tuning (SFT), quantified using normalized mean attention distance (MAD) \cite{vit}. MAD quantifies the average spatial distance between a query token and the tokens it attends to. A lower MAD indicates a local attention focus, where the model primarily attends to nearby regions; conversely, a higher MAD reflects a preference for global attention, capturing long-range dependencies across the image. Figure~\ref{fig:motiv}, indicates that models trained on salient object datasets prioritize global features, whereas those fine-tuned on camouflaged objects integrate both global and local patterns.

This deviation motivates a \textbf{natural approach} for zero-shot camouflaged object segmentation (COS), which inherently requires a fine balance between global semantic understanding and local discriminability. As shown in Figure~\ref{fig:overview}, we build upon a Masked Image Modeling (MIM) pre-trained image encoder~\cite{CLandMIM}, which effectively captures local visual features, and further augment it using SOS datasets to enhance global attention. To preserve the local representations learned through MIM, we adopt parameter-efficient fine-tuning (PEFT), ensuring that global context can be integrated without disrupting the encoder’s local sensitivity.

However, as illustrated in Figure~\ref{fig:obv}, features extracted from SOS data often remain at a coarse semantic level. To address this, we introduce a \textbf{multi-scale fine-grained alignment mechanism} guided by a Multimodal Large Language Model (M-LLM). Leveraging the semantic richness of M-LLMs, we generate descriptive captions that provide hierarchical semantic cues, allowing the model to align and refine visual embeddings across multiple scales. This alignment facilitates the capture of intricate semantic structures that are crucial for dense prediction tasks like COS. This is conceptually related to prior work in few-shot classification, such as KTN~\cite{peng2019few}, which demonstrates the utility of external semantic knowledge for classifier generalization in low-data regimes. In our setting, we adapt this principle to dense prediction by aligning M-LLM-generated semantics with multi-scale visual features.

During training, the MIM encoder outputs visual features while the M-LLM supplies corresponding caption embeddings. These two modalities are aligned within our framework to establish a synchronized understanding of both low-level and high-level cues. This synergy empowers the model to better interpret complex, ambiguous regions by grounding its predictions in fine-grained semantics. To reduce inference-time computational overhead and enable efficient deployment, we propose a novel method for training a compact query-based \textbf{codebook} that substitutes the M-LLM during inference. This learned codebook effectively captures the essential semantic priors distilled from M-LLM guidance, working jointly with the visual embeddings to produce accurate masks without relying on external semantic prompts.



To summarize, in this work we address a significant question: Is it possible to achieve zero-shot camouflaged object segmentation (COS) without manual annotations for any camouflaged object? And our contributions are as follows:

\begin{itemize}
    \item[$\bullet$] We affirmatively answer the question of zero-shot COS feasibility and introduce a new framework that supports both zero-shot and supervised learning modes for COS. This framework efficiently integrates vision and caption embeddings, utilizing a novel feature fusion module that optimizes fine-grained semantic alignment and enhances model performance. This also helps us to establish a foundational link between camouflaged object segmentation and salient object segmentation, demonstrating the beneficial interaction between these two tasks for enhancing COS.
    \item[$\bullet$] We further develop a method for training a codebook that substitutes the need for M-LLM during inference, significantly reducing the computational overhead and enabling more efficient deployment.
    \item[$\bullet$] Our approach sets new benchmarks in zero-shot camouflaged object segmentation, delivering performance on par with existing weakly-supervised methods. Furthermore, it shows robust competitiveness in supervised COS scenarios. Moreover, our method significantly outperforms challenging baselines in polyp segmentation in both zero-shot and supervised settings, showcasing its broad applicability and effectiveness across a variety of segmentation tasks.

\end{itemize}


\section{Related Work}
%
%
%
%
%

\subsection{Camouflaged Object Segmentation}
COS is a recently emerging computer vision task, that has garnered increasing attention from the community. 
To accurately localize and segment camouflaged objects, numerous deep learning-based methods have been proposed, which can be roughly divided into supervised methods and weakly-supervised methods. The former often rely on large-scale pixel-level annotated data sets, including bio-inspired methods~\cite{SINet, SINetV2}, multi-scale feature exploration~\cite{SegmMaR, FSPNet}, multi-task joint learning~\cite{sun2022bgnet, FEDER}, uncertainty-aware learning~\cite{ZoomNet}, multi-source information fusion~\cite{wu2023popnet,zoomnext}, and generative segmentation~\cite{chen2023diffusion}. The latter generally utilizes scribbles or points as supervision signals, such as contrast and relation model~\cite{CRNet} and SAM-guided learning~\cite{SAM-MFG}, which greatly reduces manual annotation. For more related work, please refer to~\cite{fan2023advances}. Nevertheless, large-scale data annotation is essential.

\subsection{Zero-Shot Learning}
Lampert \etal \cite{ZSL} firstly introduced Zero-shot learning (ZSL), a method that aims to accumulate knowledge from various datasets and identify classes that are not present in the training data. 
Recently, foundation models such as CLIP \cite{openclip, CLIP} and SAM \cite{SAM} have showcased strong zero-shot capabilities. 
CLIP leverages a large-scale dataset from web pages, demonstrating impressive zero-shot and open-vocabulary capabilities. 
Similarly, SAM exhibits strong zero-shot generalization, although it may face challenges with segmenting camouflaged objects~\cite{ji2023sam}. Notably, achieving impressive zero-shot performance often relies on carefully crafted prompts to guide the segmentation process.
Li \etal \cite{ZSCOD} proposes a novel problem called Zero-Shot Camouflaged Object Detection, the first work to detect camouflaged objects of unseen classes in a zero-shot learning setting. 
Similarly, Pang \etal \cite{OVCOS} introduces the Open-Vocabulary Camouflaged Object Segmentation (OVCOS) task, which requires the model to perform open-vocabulary segmentation of camouflaged objects and retrieve the most likely object class from a set of unseen object classes. 
However, both approaches focus on specific classes of objects, aiming to segment unseen categories. This requires the model to be trained on camouflaged objects from seen classes, meaning that it cannot eliminate the need for COS datasets.
Recently, GenSAM \cite{gensam} and MMCPF \cite{MMCPF} have utilized prompt engineering to integrate SAM \cite{SAM} and M-LLM for zero-shot camouflaged object segmentation. While these methods do not require additional post-training on COS datasets, they still encounter high computational costs during inference.
%
%
%

%
%
\subsection{Prompt Guidance}
{Prompt guidance has emerged as a critical component in multi-modal learning frameworks. In SAM \cite{SAM}, prompts such as points, bounding boxes, and masks are employed to direct segmentation tasks. In foreground segmentation, modalities like thermal and depth maps have been adopted. Thermal maps leverage heat signatures to improve robustness, as demonstrated in SOS tasks through RGB-thermal fusion approaches \cite{SODT01, SODT02}. Depth maps provide geometric cues to improve segmentation in both SOS \cite{SODD01DPANet, SODD02CPFP, SODD03} and COS \cite{CODD01}. The OVCOS framework \cite{OVCOS} further extends prompt utility by integrating depth and contour maps assisting open vocabulary COS. 
    
In contrast to prior works that employ prompts as \textbf{explicit segmentation guides}, our approach redefines the role of text prompts by utilizing them for \textbf{implicit feature learning} during training. Instead of directly localizing objects, our method leverages text prompts to learn fine-grained semantic features, thereby enabling the model to acquire transferable representations that generalize effectively across diverse segmentation tasks.}

\subsection{Parameter-Efficient Fine-Tuning (PEFT)}
PEFT reduces computational and storage needs by fine-tuning a small subset of model parameters while keeping the majority fixed. It mainly includes adapter-based techniques and Low-Rank Adaptation (LoRA).
The adapter approach, one of the techniques used in PEFT, involves inserting small modules called adapters between transformer layers \cite{PEFT}. It can be divided into two categories: In-Block adapter and Branch Adapter.
In-block adapter is a type of adapter that is inserted within a transformer block. Two common strategies for in-block adapters are serial and parallel. 
The serial strategy includes adapters such as NLP Adapter \cite{nlp_adapter,nlpadapter02} for language tasks and Mona Adapter \cite{MonaAdapter,MonaAdapter2} for vision tasks. On the other hand, AdaptFormer \cite{AdaptFormer} used the parallel strategy.
Another type of adapter is the branch adapter, which introduces adapters that branch out from specific transformer layers. Examples of branch adapters include EVP \cite{EVP, EVPV2} and SAM Adapter \cite{SAMAdapter}.
Other PEFT methods include Low-Rank Adaptation (LoRA) \cite{LoRA}, Visual Prompt Tuning (VPT) \cite{VPT}, BitFit \cite{bitfit}, \etc.

\begin{figure*}[!thb]
	\centering
	\includegraphics[width=0.99\linewidth]{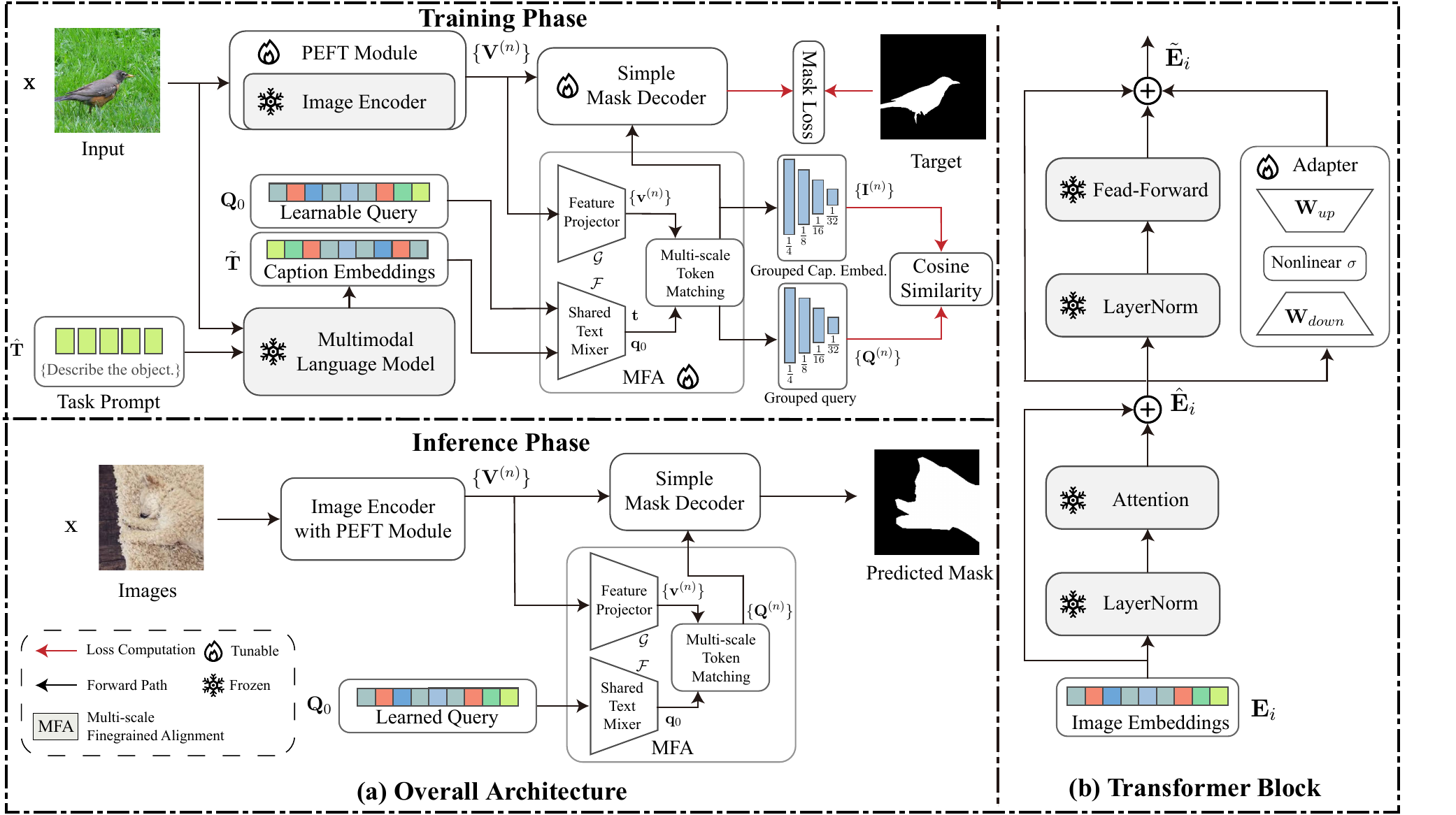}

 \caption{{\bf Overall Architecture of the Proposed Framework.} In (a), PEFT module is employed during the training phase. Specifically, only PEFT module, the MFA, the query, and the simple mask decoder are fine-tuned, while the remaining parts of the architecture are kept frozen. During inference, the M-LLM is removed and the learned query replaces the caption embeddings. (b) illustrates the implementation of PEFT using Adapter \cite{AdaptFormer} on the transformer block \cite{EVA02} within the image encoder. The bias terms is omitted in the figure.}
	\label{fig:overall}
\end{figure*}

\section{Methodology}

%

%
\subsection{Overall Architecture}

As discussed earlier, zero-shot scenarios demand effective extraction of semantic information~\cite{CLIP}, while dense prediction tasks, particularly COS models, rely on high-frequency details to capture object boundaries accurately~\cite{CLandMIM}. 
{Our empirical analysis reveals distinct attention behaviors: camouflaged object segmentation (COS) relies on both local discriminability and global semantics, whereas salient object detection primarily emphasizes global features. Motivated by this observation, we propose CAMF (Cross-modal Alignment via Multi-scale Fine-grained Fusion), a novel framework designed to bridge this gap for zero-shot COS.

CAMF integrates the SOS training dataset—which effectively provides semantic supervision and global context—with a MIM-pretrained image encoder that inherently preserves local visual details. This combination allows the model to benefit from both semantic richness and multi-scale visual fidelity. However, full fine-tuning may disrupt the local features learned during MIM pre-training due to catastrophic forgetting. To avoid this, we adopt parameter-efficient fine-tuning (PEFT) strategies, ensuring stable integration of new knowledge while maintaining local feature integrity. This design is aligned with recent advances in fine-tuning strategies such as Singular Value Fine-tuning (SVF)~\cite{sun2022singular}, which demonstrate that updating a small subset of backbone parameters—e.g., only singular values—can effectively adapt models to novel tasks while maintaining the integrity of pre-trained features.

Further, although SOS data promotes global pattern learning, its representations tend to be semantically coarse. To address this, we introduce a Multi-scale Fine-grained Alignment (MFA) module, which incorporates caption embeddings generated by a Multimodal Large Language Model (M-LLM). Inspired by the success of~\cite{SPARC}, MFA refines hierarchical semantics through cross-modal alignment, enabling the model to capture intricate object-level details essential for accurate dense prediction in complex COS scenes. Such a design echoes recent advances in multi-modal fusion~\cite{tang2024divide}, where modality-specific and complementary information are decoupled to enhance robustness. Similarly, our framework aims to extract fine-grained complementary semantics from M-LLM captions to augment visual embeddings for complex COS scenes.

To support efficient inference, we further propose a codebook-based decoding strategy that eliminates the need for M-LLM at test time. This codebook, trained during alignment, replaces caption embeddings with learned queries, thereby preserving semantic guidance while significantly reducing inference-time computational overhead. }

\subsection{Image Encoder with PEFT}
{To minimize the effects of catastrophic forgetting, we employ adapter as PEFT module. Specifically, we introduce adapters within the MLP modules to avoid disturbing local features learned by the attention mechanisms. Additionally, we utilize an EVA-pretrained model~\cite{EVA01,EVA02}, which benefits from both MIM and contrastive learning (CL) pre-training, to enhance the convergence behavior of PEFT. For subsequent multiscale  fine-grained alignment, we introduce a simple module to produce multiscale visual features.}

%
%

%

%
%
\noindent {\bf Parameter-Efficient Finetuning.}
%
%
%
During training, given an image $\mathbf{X}\in \mathbb{R}^{H\times W \times 3}$, the patch embedding operation divide it into patches and projects these into image embeddings $\mathbf{E}_1 \in \mathbb{R}^{L\times d_v}$ before passing them into the transformer block of the model.  
Originally, the transformer block can be formulated as:
\begin{align}
    {\bf \hat{E}}_i &= \textrm{Attention}( \textrm{LN}( {\bf E}_i ) ) + {\bf E}_i, \\ 
    {\bf \tilde{E}}_i &= \textrm{FFN}( \textrm{LN}( {\bf \hat{E}}_i ) ) + {\bf \hat{E}}_i, \label{eq:ffn}
\end{align}
where $i$ denotes the $i^{\text{th}}$ transformer block, $\mathbf{E}_i$ is the image embedding from the $(i-1)^{\rm th}$ block, such that $\mathbf{E}_i={\bf \tilde{E}}_{i-1}$. 
In this work, We used a parallel adapter in \cite{AdaptFormer,PEFT} to perform PEFT. Given a matrix $\mathbf{Z}\in \mathbb{R}^{L \times d_v}$, the adapter is a low-rank MLP that can be denoted as 
\begin{equation}
    \text{Adapter}\left( \mathbf{Z} \right) = \sigma \left( \mathbf{Z} {\bf W}_{down} + \mathbf{b}_1 \right) {\bf W}_{up} + \mathbf{b}_2, \label{eq:adapter}
\end{equation}
where $\sigma$ represents any non-linear activation function, ${\bf W}_{down}\in \mathbb{R}^{d_v\times d_{lr}}$ and ${\bf W}_{up} \in \mathbb{R}^{d_{lr}\times d_v}$ forms a low rank form with $d_{lr} \ll d_v$. In this work, we use ReLU \cite{relu} as activation function. 
%
%
Unlike \cite{AdaptFormer,PEFT}, we have removed the layer normalization and added bias terms for the two linear layers. This modification aims to enable the adapter to learn shape-biased features better at the channel level.
Then the adapter represented in Eq.~\ref{eq:adapter} is injected into the branch of FFN within the transformer blocks. 
The Eq.~\ref{eq:ffn} can be rewritten as 
\begin{equation}
	{\bf \tilde{E}}_i = a_i \cdot \text{Adapter}\left({\bf \hat{E}}_i \right) + \textrm{FFN}\left( \textrm{LN}\left( {\bf \hat{E}}_i \right) \right) + {\bf \hat{E}}_i,
 \label{equ:new}
\end{equation} where $a_i \in \mathbb{R}$ is a learnable scale factor.
\begin{figure*}[t]
	\centering
	\includegraphics[width=0.99\linewidth]{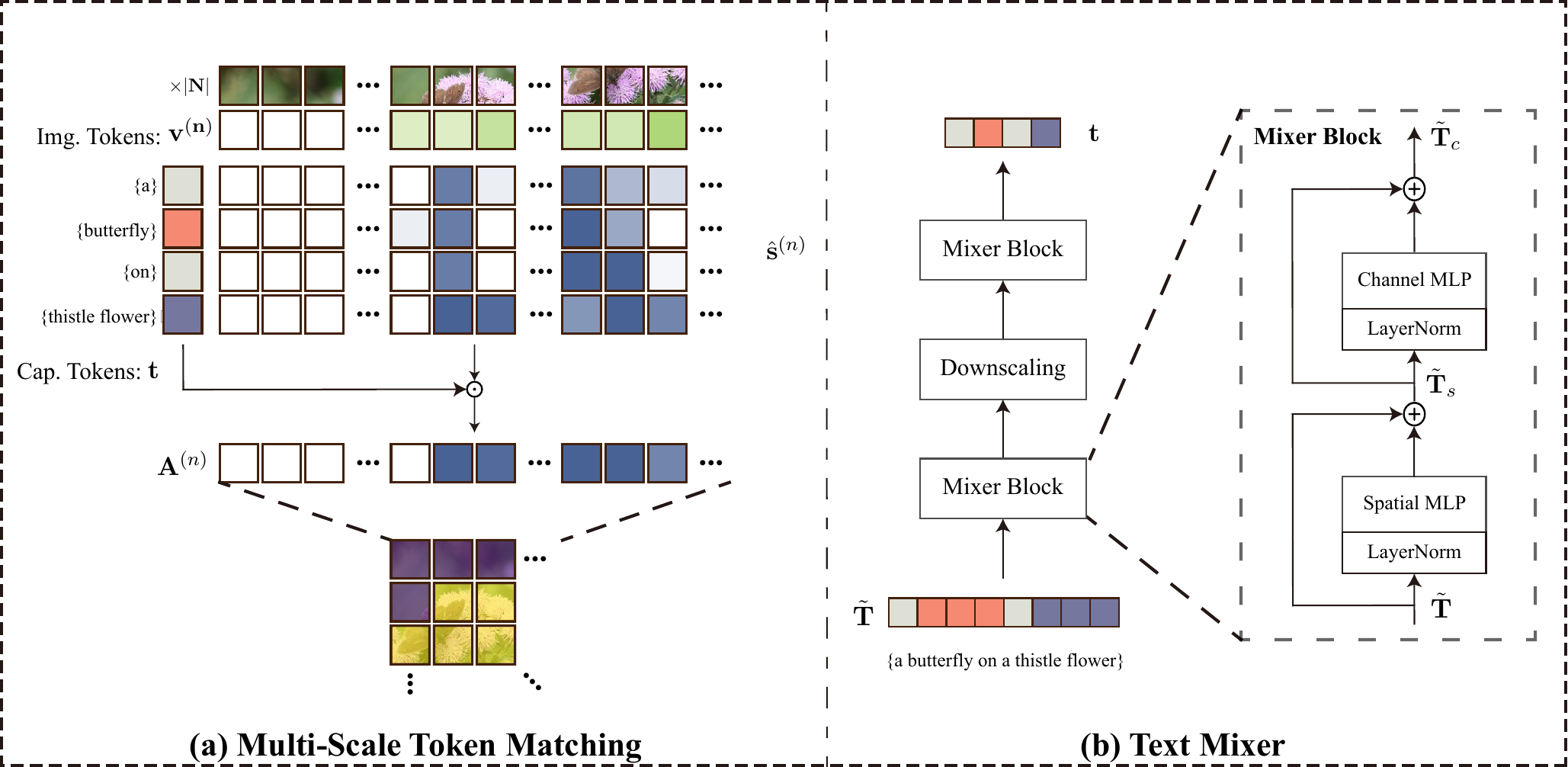}
	\caption{{\bf Multi-scale Token Match Operation and the structure of Text Mixer.} The left figure highlights how the multi-scale vision tokens and caption tokens are aligned using TM. The right figure provides a closer look at the text mixer, which is composed of two mixer blocks and a down-scaling module.
	}
	\label{fig:MFA}
\end{figure*}

\noindent {\bf Multi-scale Transform.} 
%
%
Motivated from \cite{ViTDet}, we utilize simple feature pyramids (SFP) to generate multi-scale features. More specifically, the last feature map of image encoder $\mathbf{\tilde{E}}_N \in \mathbb{R}^{L_v\times d_v }$, where $L_v=\frac{HW}{16^2}$, is fed into SFP to produce multi-scale feature maps $\{{\bf V}^{(n)}\}$, where $n\in \mathbf{N} $ and $\mathbf{N} = \{4,8,16,32\}$, at scales $\{ \frac{HW}{32^2},\frac{HW}{16^2},\frac{HW}{8^2}, \frac{HW}{4^2}  \}$.
\subsection{Caption Embeddings from M-LLM}
%
%
{To provide cross-modal information for fine-grained alignment and learn rich semantic content, we employ a M-LLM to generate textual embeddings as captions. The M-LLM takes both an image and a prompt as inputs, and processes them to generate a concise caption embedding that captures the necessary information relevant to the image.}

%

%
Given an image $\mathbf{X}$ and a task prompt $\mathbf{T}$, $\mathbf{X}$ and $\mathbf{T}$ are sent to the M-LLM to generate a caption embedding $\tilde{\mathbf{T}} \in \mathbb{R}^{L_t\times d_t}$.
\begin{equation}
 \tilde{\mathbf{T}} = \text{M-LLM}\left(\mathbf{X}, \mathbf{T}\right).
\end{equation}
Different prompts offer varied inputs for the language model, capturing different task-related information. Each prompt may lead to different performance outcomes for the model. In Section~\ref{sec:abl}, we include an ablation study on the choice of prompts in both zero-shot and supervised learning.

\subsection{Multi-Scale Fine-Grained Alignment}

Multi-Scale Fine-Grained Alignment (MFA) is a key component of our framework, designed to align the caption embeddings generated by the M-LLM with the image embeddings from the fine-tuned encoder. This process enhances the PEFT module’s ability to learn fine-grained semantic features. Additionally, we initialize a learnable query to build a codebook, which significantly improves efficiency during inference. By using the codebook as a substitute for the M-LLM, the model bypasses the computationally expensive process of generating caption embeddings, resulting in faster inference times without sacrificing accuracy. This streamlined approach allows for more efficient deployment, particularly in scenarios where low-latency responses are critical.

%
%

As shown in Figure~\ref{fig:overall}, the MFA module consists of three key components: a feature projector, a multi-scale token matching operation (Figure~\ref{fig:MFA} (a)), and a text mixer (Figure~\ref{fig:MFA} (b)). 
{The MFA takes both the caption embeddings ${\bf \tilde{T}}$, and the multi-scale image embeddings $\{ {\bf V}^{(n)} \}$, as inputs, producing image-grouped caption embeddings $\{ {\bf I}^{(n)} \}$, that are compatible with the mask decoder. Additionally, the learnable query ${\bf Q}_0$ is mapped alongside $\{ {\bf V}^{(n)} \}$ to generate image-grouped queries $\{ {\bf Q}^{(n)} \}$.}

\noindent {\bf Multi-scale Token Matching.} 
To achieve the fine-grained alignment and train a codebook, we propose a simple attention-like mechanism that aligns caption tokens and query tokens with image tokens generated by the text mixer and the feature projector. This mechanism, illustrated in Figure~\ref{fig:MFA}~(a), is inspired by techniques from \cite{SPARC}, with several modifications implemented to enhance its effectiveness.
%

%

%

During this process, the multi-scale image embeddings $\{ {\bf V}^{(n)} \}$ are first passed through a feature projector $\mathcal{G}:\mathbb{R}^{L_v^{(n)}\times d_v} \rightarrow \mathbb{R}^{L_v^{(n)}\times {d} } $, which generates multi-scale image tokens $\{\mathbf{v}^{(n)}\}$. All image embeddings at each scale utilize the same image projector.
%
Similarly, the caption embedding ${\bf \tilde{T}}$ and the query $\mathbf{Q}_0$ are processed by a shared text mixer $\mathcal{F}:\mathbb{R}^{L_t\times d_t} \rightarrow \mathbb{R}^{L_t / 2\times {d} }$ to compress the embeddings. Therefore, we have the following notations:
\begin{align}
    \mathbf{v}^{(n)} &= \mathcal{G}\left(\mathbf{V}^{(n)} \right), \\
    \mathbf{t} &= \mathcal{F}\left(\tilde{\mathbf{T}} \right), \\
    \mathbf{q}_0 &= \mathcal{F}\left(\mathbf{Q}_0 \right).
\end{align}

For simplicity, we denote the sequence tokens from the mixer is $\mathbf{u}$, where $\mathbf{u}\in \mathbb{R}^{L_t / 2\times d}$ and $\mathbf{u} \in \{ \mathbf{t}, \mathbf{q}_0\}$. Therefore, the entire token matching process can be represented as:
\begin{equation}
    \mathbf{A}^{(n)} = \text{TM}\left( \mathbf{v}^{(n)}, \mathbf{u} \right), \label{eq:tm}
\end{equation}
where $\text{TM}(*)$ denotes token matching operation.
Specifically, The first step is to calculate the similarity and attain the alignment weights from the similarity using min-max normalization along the row.
\begin{equation}
	\mathbf{s}^{(n)} = \text{Normalize}_{L_t}\left(\mathbf{v}^{(n)} \mathbf{u}^\top \right).
\end{equation}
The resulting $\mathbf{s}^{(n)}$ is a normalized similarity matrix with a shape of $L_v \times \frac{L_t}{2}$. Then, the alignment weights are sparsified using a simple threshold strategy. We denote ${\bf s}^{(n)}=[s_{ij}^{(n)}]$ and $\hat{\bf s}^{(n)}=[\hat{s}_{ij}^{(n)}]$, where $1\le i\le L_v^{(n)}$ and $1\le j\le \frac{L_t}{2}$.
%
%
\begin{equation}
    \hat{s}_{ij}^{(n)} = 
    \left\{
        \begin{array}{rcl}
		s_{ij}^{(n)},       & s_{ij}^{(n)} \ge \frac{1}{L_v^{(n)}}\\
		0,     & \quad {\rm otherwise}.\\
       \end{array} 
    \right. 
\end{equation}
Every element in ${\bf s}^{(n)}$ undergoes a thresholding process. This thresholding is based on the threshold value $1/{L_v^{(n)}}$. If an element is equal to or greater than the threshold value, its value remains; otherwise, it is assigned a value of 0.
Then, we get the image-grouped tokens by
\begin{equation}
	\mathbf{A}^{(n)} = \text{Softmax}_{L_t}\left(\hat{\bf s}^{(n)} \right) \ \mathbf{u},
\end{equation}
where $\text{Softmax}_{L_t}(*)$ denotes the softmax operation along the row. 

This operation is applied to caption tokens and query tokens from the mixer separately. 
Subsequently, by replacing $\mathbf{u}$ with caption and query tokens $\mathbf{t}$ and $\mathbf{q}_0$ respectively, the multi-scale TM operation produces the multi-scale image-grouped caption embeddings $\{ {\bf I}^{(n)} \}$ and the image-grouped queries $\{ {\bf Q}^{(n)} \}$. This process can be expressed using Eq.~\ref{eq:tm} as follows:
%
%
%
\begin{align}
    {\bf I}^{(n)} &= \textrm{TM}\left( \mathcal{G}\left( {\bf V}^{(n)} \right), \mathcal{F}\left( {\bf \tilde{T}} \right) \right), \\
    {\bf Q}^{(n)} &= \textrm{TM}\left( \mathcal{G}\left( {\bf V}^{(n)} \right), \mathcal{F}\left( {\bf Q}_0 \right) \right).
\end{align}
%
%

%

%

%

%
For caption tokens, this operation produces image-grouped caption embeddings that are aligned with the image embeddings. For example, if the caption mentions a ``poodle'', the corresponding patches in the image-grouped caption embeddings representing the ``poodle'' will be activated, while the remaining patches will be deactivated. Furthermore, by employing multi-scale TM, we obtain the image-grouped query $\{ \mathbf{Q}^{(n)} \}$ for the codebook learning.

%
%
%

%

\noindent {\bf Text Mixer and Feature Projector.} 
Text mixer and image projector are designed to learn non-linear transformations. Thus, we can learn a more expressive and flexible mapping between the visual representations and the aligned feature space, beyond what a simple linear projection could achieve. 
%
%
The text mixer $\mathcal{F}$ applies both spatial and channel transformations to the caption sequence, allowing for more fine-grained processing of the textual representations. 
MLP based mixer, are generally more computationally efficient than attention-based architecture.
This text mixer is composed of two mixer block and a linear down-scaling in the middle. Each block contains a spatial mixer and a channel mixer based on MLP. The mixer block can be formulated as follows:
\begin{align}
    \tilde{\bf T}_s &= {\rm MLP}\left( {\rm LN}\left(\tilde{\bf T} \right)^{\top} \right) + \tilde{\bf T}, \\
    \tilde{\bf T}_c &= {\rm MLP}\left( {\rm LN}\left(\tilde{\bf T}_s\right) \right) +  \tilde{\bf T}_s.
\end{align}
Then, $\hat{\bf T}_c$ is down-scaled by a learnable matrix $\mathbf{W}_{/ 2}\in \mathbb{R}^{ \frac{L_t}{2}\times d }$ and another mixer block is apply on it to get the final $\mathbf{t}$.
For image embeddings, the feature projector is a simple channel MLP which maps $\mathbb{R}^{L_v^{(n)}\times d_v}$ to $\mathbb{R}^{ 
L_v^{(n)}\times d }$ allowing us to adjust the channel dimensionality of the image embeddings to match the channel dimensionality of the caption embeddings.

\subsection{ Mask Decoder}

\begin{figure}[t]
	\centering
	\includegraphics[width=0.90\linewidth]{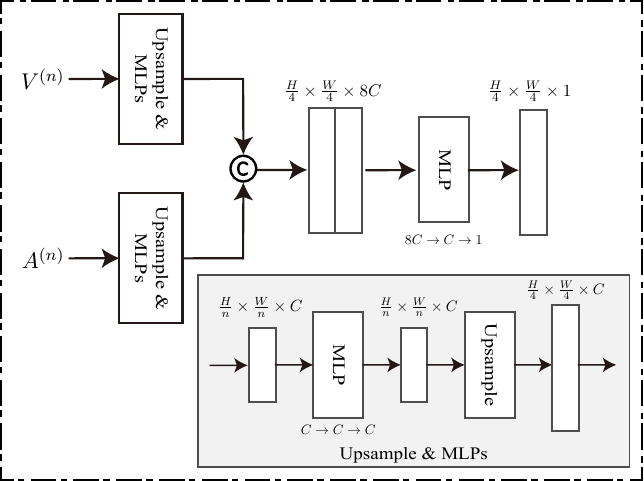}
	\caption{{\bf The structure of simple mask decoder.} We use several MLPs and upsampling modules to maintain a simple design. 
	}
	\label{fig:mask_dec}
\end{figure}

We utilize an all-MLP decoder to maintain a straightforward design. As illustrated in Figure~\ref{fig:mask_dec}, the mask decoder comprises nine MLP layers in total. It takes both multi-scale image embeddings $\{ \mathbf{V}^{(n)} \}$ and grouped tokens $\{ \mathbf{A}^{(n)} \}$ as inputs. These two representations are then fused using a simple concatenation operation. The fused representation is processed through an MLP to produce the final output mask logits.

\subsection{Inference with Learned Query}
%
During the inference phase, the M-LLM is not used. Instead, we replace the caption embeddings with the learned query $\mathbf{Q}_0$. 
In the training phase, $\mathbf{I}^{(n)}$ serves as $\mathbf{A}^{(n)}$ for the input to the simple mask decoder. During inference, since the M-LLM is replaced, $\mathbf{Q}^{(n)}$ is used as $\mathbf{A}^{(n)}$ for the input to mask decoder.
Surprisingly, in Section \ref{sec:abl}, we demonstrate that this approach outperforms using caption embedding from the M-LLM during inference. 
%

%

\subsection{Loss Function}
\label{sec:loss_func}
The total loss can be represented as a weighted sum of mask loss $\mathcal{L}_{mask}$ and query loss $\mathcal{L}_q$. The mask loss is used to regularize the segmentation mask generation. The query loss is used to regularize the query to learn a codebook. 
\begin{equation}
    \mathcal{L} = \mathcal{L}_{mask} + \lambda \mathcal{L}_q,
\end{equation}
with $\lambda > 0$.  
\noindent{\bf Mask Loss.}
We use BCE, DICE, and UAL \cite{zoomnext} for mask loss. The BCE loss function is a fundamental object function in various binary image segmentation tasks. The DICE loss and UAL loss are used to increase the confidence.
\begin{equation}
    \mathcal{L}_{mask} = \lambda _{bce}\mathcal{L}_{bce} + \lambda _{dice}\mathcal{L}_{dice} + \lambda _{ual}\mathcal{L}_{ual}.
\end{equation}

\noindent{\bf Query Loss.}
Cosine similarity is utilized to learn the query effectively. For each scale of the grouped caption embeddings and their corresponding queries, we compute the cosine similarity to evaluate their relationship.

\begin{equation}
    \mathcal{L}_q = 1 - \frac{1}{|\mathbf{N}|} \sum_{n\in \mathbf{N} } \frac{ {\bf I}^{(n)}\cdot {\bf Q}^{(n)} }{ ||{\bf I}^{(n)}||_2\cdot ||{\bf Q}^{(n)}||_2 }.
\end{equation}

\section{Experiments}
\label{sec:exp}

\subsection{Experiment Setting }
\noindent {\bf Zero-shot Setting.}
To perform zero-shot COS, we used the salient object segmentation dataset DUTS \cite{DUTS} during the training phase.
During the evaluation phase, we employed four widely used COS datasets to thoroughly assess the model's zero-shot capabilities: the test sets of CAMO \cite{CAMO}, COD10K \cite{SINetV2}, CHAMELEON \cite{CHAMELEON}, and NC4K \cite{NC4K}.
In the zero-shot learning setting for COS, models were trained for 20 epochs with a linear warm-up over the first 6 epochs on DUTS, using a batch size of 10.

\noindent {\bf Supervised Learning Setting.}
In supervised camouflaged object segmentation (COS), we used a combined training set from CAMO and COD10K, while the remaining COS datasets served as the test datasets.
Specifically, CAMO \cite{CAMO} provided 1,000 training images and 250 testing images featuring diverse camouflaged objects. COD10K \cite{SINetV2} included 3,040 training samples and 2,026 test samples. CHAMELEON \cite{CHAMELEON} and NC4K \cite{NC4K} contained 76 test samples and 4,121 samples, respectively.
For supervised learning in COS, we trained the models for 50 epochs with a linear warm-up of 10 epochs.

\noindent {\bf Implementation Details.}
%
%
%
%
%
Our method was implemented using the PyTorch framework and the experiments were conducted on a GeForce RTX4060Ti GPU. 
In our implementation, we utilized BLIP2 \cite{blip2} offline to save GPU memory, specifically employing the {\it blip2-opt} model as the M-LLM component. The standalone image encoder was a pre-trained {\it EVA02-L} \cite{EVA02} model, while the remaining modules were randomly initialized. To optimize the attention mechanism and further reduce GPU memory usage, we utilized the FlashAttention operator \cite{dao2022flashattention}.
We employed the AdamW \cite{adamw} optimizer for all experiments, with an initial learning rate of $1.5\times {10}^{-4}$ and a cosine decay scheduler. During the training phase, we performed data augmentation by resizing input images to 384$\times$384 pixels and applying random horizontal flips. 
%

%

%

\noindent{\bf Loss Weights.} 
The loss function employed for training was a combination of mask loss and query loss see in Seection~\ref{sec:loss_func}. 
We employed a straightforward approach for combining loss weights. In zero-shot, for the mask loss $\mathcal{L}_{mask}$, we set $\lambda_{bce}=1$, and $\lambda_{dice}=0.5$. The UAL loss was not included in this setting. The weight for the query loss was set to $\lambda = 0.5$. In supervised training, we maintained the same values for $\lambda_{bce}$, $\lambda_{dice}$, and $\lambda$. To enhance the confidence of the segmentation map, we included an additional UAL loss \cite{zoomnext}. Instead of using a dynamic loss weight for the UAL loss as described in \cite{zoomnext}, we fixed the weight at $\lambda_{ual}=1$.

\noindent {\bf Evaluation Metrics.}
To thoroughly evaluate the model's performance, we employed four widely-used metrics: weighted F-measure ($F^w_\beta$) \cite{wfm}, mean absolute error (MAE), S-measure ($S_\alpha$) \cite{sMeasure}, and mean E-measure ($E_\phi$) \cite{Emeasure, Emeasure2}. 
%

%
\begin{figure*}[th]
	\centering
	\includegraphics[width=0.81\linewidth]{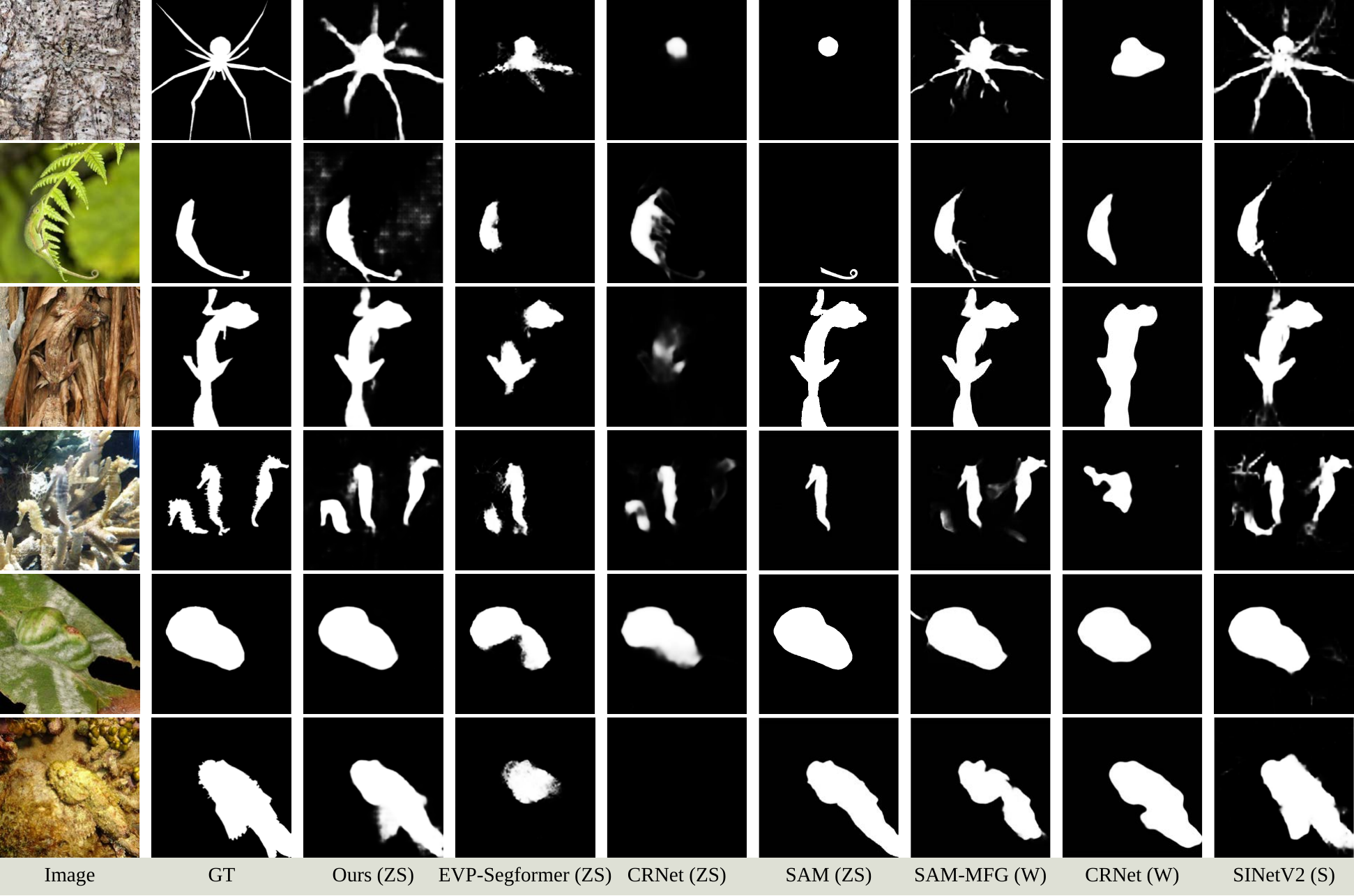}
	\caption{{\bf Comparison with Other Methods under Zero-shot, Weakly Supervised, and supervised Learning Setting.}
	We compare our proposed method with existing approaches under zero-shot (ZS), weakly supervised (W), and supervised (S) learning settings. The results are presented for the COD10K \cite{SINetV2}, and CAMO \cite{CAMO} datasets. 
	}
	\label{fig:zs_visual}
\end{figure*}
%
%

\subsection{Zero-Shot Results}
{\noindent{\bf Comparison Methods.}
In Table~\ref{tab:results}, we present a comparative evaluation of our method against various approaches across four datasets. The compared methods include weakly supervised models such as CRNet~\cite{CRNet} and SAM-MFG~\cite{SAM-MFG}, as well as M-LLM-based methods including GenSAM~\cite{gensam} and MMCPF~\cite{MMCPF}. {The unsupervised version of CDP \cite{CDP} shares a similar training paradigm with SAM-MFG which leverages pseudo-labels. Therefore, we group CDP under the WS setting. For a fair comparison, we also report the performance of several existing methods under our zero-shot setting, where training is conducted solely on the SOS dataset. }

\noindent{\bf Comparison Settings.}
%
To ensure a rigorous evaluation under our zero-shot setting, we established multiple comparison baselines under controlled conditions.
First, to ensure that the performance gains of our method are not solely attributable to the backbone architecture, we introduced EVP-EVA02-L~\cite{EVP,EVA02}, which adopts the same backbone (EVA02-L~\cite{EVA02}) as our proposed method and is trained under identical hyperparameters. Additionally, we included EVP-SegFormer (trained on the SOS dataset in its original paper) \cite{EVP,SegFormer} to further validate the robustness of our approach.
For assessing zero-shot capability of SOS methods on COS tasks, we selected the classical VST model with its original DUTS-trained weights \cite{VST}, establishing a baseline for cross-task generalization.
Furthermore, to evaluate the generalization of both supervised and weakly supervised methods in a zero-shot COS setting, we retrained SINet-V2 (supervised) \cite{SINetV2} and CRNet (weakly-supervised) \cite{CRNet} using the same hyperparameters and training protocols as our method.

The performance metrics for SCWSSOD~\cite{SCWSSOD} and TEL~\cite{TEL} are obtained from~\cite{SAM-MFG}, while SAM's results are sourced from~\cite{gensam} and SAM2's from~\cite{sam2zscod}. All other baseline results are taken directly from their original publications. For metrics not reported in the original papers, we have reproduced the missing results according to their published specifications.}

{\noindent{\bf Discussion.}
Our method demonstrates strong performance in zero-shot COS task, showing the highest weighted F-measure compared to other methods under zero-shot setting in four datasets. Notably, a comprehensive analysis of Table~\ref{tab:results} reveals an implicit trend: while certain methods excel on individual datasets (\eg CRNet \cite{CRNet} achieving a weighted F-measure of 0.744 on the CHAMELEON \cite{CHAMELEON} dataset), their performance diminishes on other datasets (\eg CRNet \cite{CRNet} achieving a weighted F-measure of only 0.576 on the COD10K \cite{SINetV2} dataset). In contrast, our method consistently maintains strong performance across all four datasets, with weighted F-measures exceeding 0.710, MAE values below 0.107, S-measure values surpassing 0.775, and mean E-measure values exceeding 0.800. 
In Figure~\ref{fig:zs_visual}, we present visual comparisons between our method and other COS approaches under zero-shot setting. The results demonstrate that our proposed method outperforms the alternative methods in terms of mask prediction for the majority of samples.}

%

%
%
%

\begin{figure}[t]
	\centering
	\includegraphics[width=0.99\linewidth]{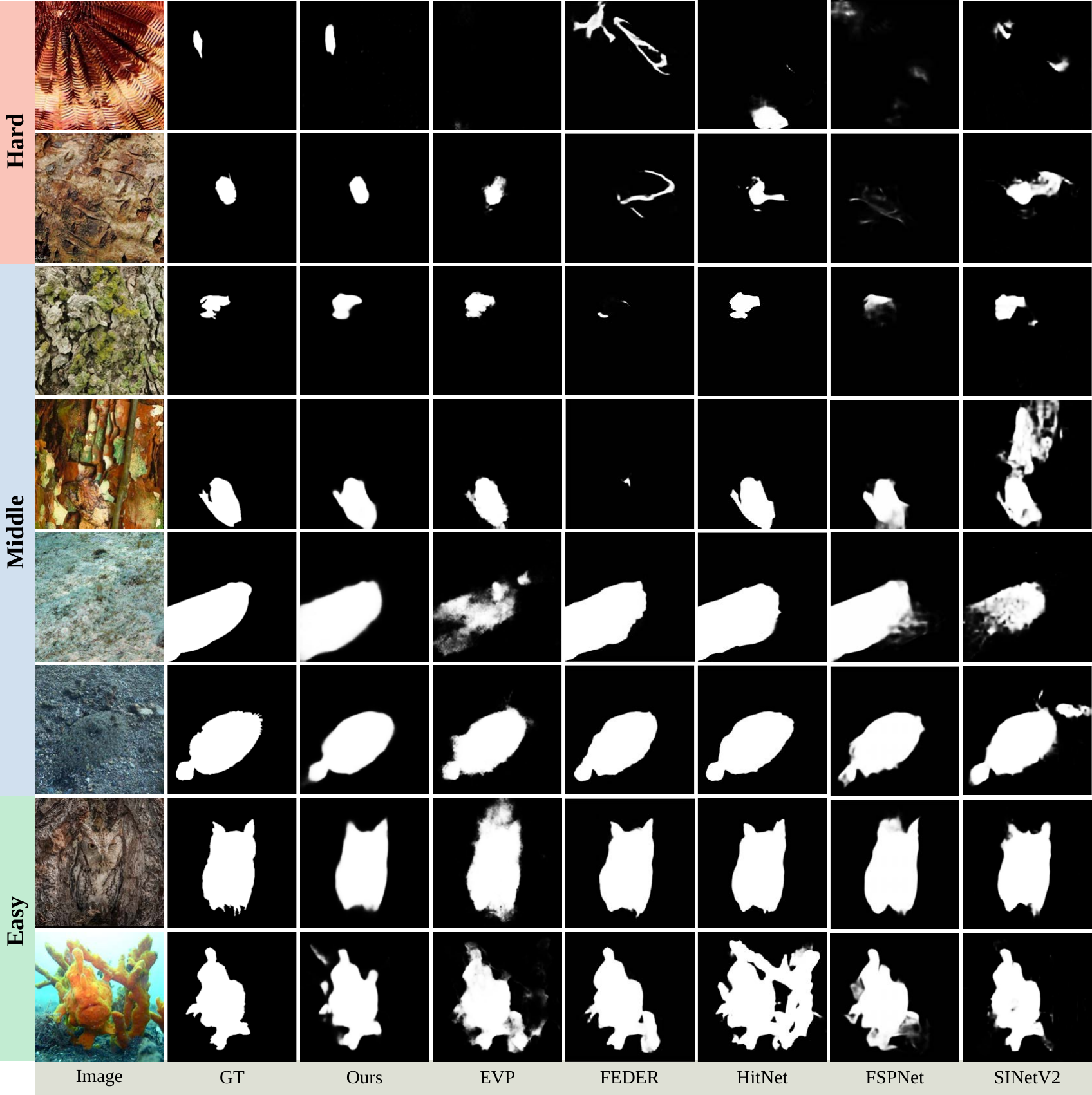}
	\caption{
    {\bf Visualization of the predicted masks of our methods and others in the supervised learning setting.} The figure presents results for hard, medium, and easy examples. Each row shows the original image, ground truth (GT), and outputs from various methods: Ours, EVP, FEDER, HitNet, FSPNet, and SINetV2. 
	}
	\label{fig:s_visual}
\end{figure}

\subsection{Supervised Learning Results}

\begin{table*}[t]
    \centering
    \caption{Quantitative results of zero-shot (ZS) and supervised camouflaged object segmentation methods. Several weakly-supervised methods (WS) are also included for comparison in the zero-shot setting. The best three results are highlighted in {\color[HTML]{FE0000} red}, {\color[HTML]{3166FF} blue}, and {\color[HTML]{32CB00} green}, respectively.}
    \setlength{\tabcolsep}{2.5pt}
    \renewcommand\arraystretch{1.2}
    
    \resizebox{\linewidth}{!}{

    \begin{tabular}{@{}l|r|c|c|cccc|cccc|cccc|cccc@{}}
    \toprule
    \rowcolor{lightgray}
    \multicolumn{20}{c}{\textbf{Zero-Shot (ZS) and Weakly-Supervised (WS) Setting }} \\ \midrule
    \multirow{2}{*}{Methods}     & \multirow{2}{*}{Pub.~~~} & \multirow{2}{*}{Sup.} & \multirow{2}{*}{Data} & \multicolumn{4}{c|}{CAMO}                                                                   & \multicolumn{4}{c|}{NC4K}                                                                   & \multicolumn{4}{c|}{COD10K}                                                                 & \multicolumn{4}{c}{CHAMELEON}                                                               \\ \cline{5-20} 
                                 &                          &                       &                       & $F_{\beta}^w \uparrow$ & $\text{MAE}\downarrow$ & $S_{\alpha}\uparrow$ & $E_{\phi}\uparrow$ & $F_{\beta}^w \uparrow$ & $\text{MAE}\downarrow$ & $S_{\alpha}\uparrow$ & $E_{\phi}\uparrow$ & $F_{\beta}^w \uparrow$ & $\text{MAE}\downarrow$ & $S_{\alpha}\uparrow$ & $E_{\phi}\uparrow$ & $F_{\beta}^w \uparrow$ & $\text{MAE}\downarrow$ & $S_{\alpha}\uparrow$ & $E_{\phi}\uparrow$ \\ \midrule
    SCWSSOD \cite{SCWSSOD,SAM-MFG}       & AAAI21                   & WS                     & COD10K                & .618                   & .102                   & .713                 & .815               & -                      & -                      & -                    & -                  & .546                   & .049                   & .710                 & .805               & .714                   & .053                   & .792                 & .881               \\
    TEL \cite{TEL,SAM-MFG}               & CVPR22                   & WS                     & COD10K                & .635                   & .133                   & .645                 & .674               & .711                   & .085                   & .766                 & .795               & .604                   & .063                   & .727                 & .803               & .645                   & .094                   & .746                 & .751               \\
    CRNet \cite{CRNet}           & AAAI23                   & WS                     & COD10K                & .641                   & .092                   & .735                 & .815               & .741                   & .080                   & .758                 & .796               & .576                   & .049                   & .733                 & .805               & .744                   & .046                   & .818                 & .897               \\
    SAM-MFG \cite{SAM-MFG}       & NeurIPS23                & WS                     & COD10K                & .679                   & .102                   & .718                 & .757               & .799                   & .057                   & .813                 & .859               & .698                   & .039                   & .790                 & .856               & .728                   & .056                   & .805                 & .868               \\ 
    CDP \cite{CDP}       & ESWA25                & WS                     & \begin{tabular}[c]{@{}c@{}}DUTS+CAMO\\ +COD10K\end{tabular}                & .678                   & .113                   & .723                 & .798               & .717                   & .075                   & .769                 & .842               & .565                   & .074                   & .704                 & .780               & .654                   & .081                   & .729                 & .833       \\ \midrule
    SAM \cite{SAM,gensam}               & ICCV23                   & ZS                    & SA-1B                 & .597                   & .160                   & .643                 & .639               & .696                   & .078                   & .767                 & .776               & .673                   & .093                   & .730                 & .737               & .595                   & .207                   & .635                 & .647               \\
    SAM2 \cite{ravi2024sam2,sam2zscod}               & arXiv24                   & ZS                    & SA-V                 & .184                   & .236                   & .444                 & .401               & .251                   & .186                   & .512                 & .482               & .271                   & .134                   & .549                 & .521               & -                   & -                   & -                 & -               \\
    GenSAM \cite{gensam}         & AAAI24                   & ZS                    & SA-1B                 & .659                   & .113                   & .719                 & {\color[HTML]{3166FF} .775}               & .665                   & .087                   & .754                 & .781               & {\color[HTML]{3166FF} .681}                   & {\color[HTML]{009901} .067}                   & {\color[HTML]{009901} .775}                 & {\color[HTML]{CB0000} .838}               & {\color[HTML]{3166FF} .680}                   & {\color[HTML]{009901} .090}                   & {\color[HTML]{3166FF} .764}                 & {\color[HTML]{3166FF} .807}               \\
    MMCPF \cite{MMCPF}           & MM24                     & ZS                    & SA-1B                 & {\color[HTML]{3166FF} .683}                   & {\color[HTML]{CB0000} .101}                   & {\color[HTML]{3166FF} .751}                 & {\color[HTML]{009901} .772}               & .686                   & .079                   & .769                 & .796               & .596                   & {\color[HTML]{3166FF} .062}                   & .769                 & {\color[HTML]{009901} .826}               & -                      & -                      & -                    & -                  \\ \midrule
    VST \cite{VST}               & ICCV21                   & ZS                    & DUTS                  & .563                   & .154                   & .651                 & .683               & .685                   & .091                   & .752                 & .790               & .544                   & .087                   & .676                 & .732               & .548                   & .113                   & .659                 & .686               \\
    SINet-V2 \cite{SINetV2}      & TPAMI22                  & ZS                    & DUTS                  & .437                   & .188                   & .583                 & .575               & .551                   & .127                   & .667                 & .668               & .404                   & .103                   & .604                 & .598               & .291                   & .165                   & .515                 & .461               \\
    CRNet \cite{CRNet}           & AAAI23                   & ZS                    & DUTS                  & .445                   & .156                   & .593                 & .575               & .599                   & .096                   & .694                 & .697               & .443                   & .079                   & .611                 & .590               & .377                   & .129                   & .562                 & .509               \\
    EVP-Segformer \cite{EVP,SegFormer}     & CVPR23                   & ZS                    & DUTS                  & .637                   & .137                   & .701                 & .733              & {\color[HTML]{009901} .756}                   & {\color[HTML]{009901} .071}                   & {\color[HTML]{009901} .788}                 & {\color[HTML]{009901} .822}               & { .652}                   & {\color[HTML]{3166FF} .062}                   & { .716}                 & { .764}               & { .642}                   & { .091}                   & {\color[HTML]{009901} .713}                 & { .713}               \\
    EVP-EVA02-L \cite{EVP,EVA02} & \multicolumn{1}{c|}{-}   & ZS                    & DUTS                  & {\color[HTML]{009901} .671}                   & {\color[HTML]{009901} .125}                   & {\color[HTML]{009901} .750}                 & {\color[HTML]{3166FF} .775}               & {\color[HTML]{3166FF} .783}                   & {\color[HTML]{3166FF} .069}                   & {\color[HTML]{3166FF} .837}                 & {\color[HTML]{3166FF} .862}               & {\color[HTML]{009901} .674}                   & { .068}                   & {\color[HTML]{3166FF} .781}                 & { .816}               & {\color[HTML]{009901} .663}                   & {\color[HTML]{3166FF} .089}                   & {\color[HTML]{009901} .760}                 & {\color[HTML]{009901} .768}               \\ \midrule
    \rowcolor{tablecl} CAMF      & \multicolumn{1}{c|}{-}   & ZS                    & DUTS                  & {\color[HTML]{CB0000} .729}          & {\color[HTML]{3166FF} .105}          & {\color[HTML]{CB0000} .788}        & {\color[HTML]{CB0000} .814}      & {\color[HTML]{CB0000} .818}          & {\color[HTML]{CB0000} .054}          & {\color[HTML]{CB0000} .861}        & {\color[HTML]{CB0000} .882}      & {\color[HTML]{CB0000} .717}          & {\color[HTML]{CB0000} .055}          & {\color[HTML]{CB0000} .808}        & {\color[HTML]{3166FF} .832}      & {\color[HTML]{CB0000} .720}          & {\color[HTML]{CB0000} .067}          & {\color[HTML]{CB0000} .805}        & {\color[HTML]{CB0000} .817}      \\ \bottomrule
    \end{tabular}
    }

    \vspace{2em} 

    \resizebox{\linewidth}{!}{
    \begin{tabular}{@{}l|r|c|cccc|cccc|cccc|cccc@{}}
    \toprule
        \rowcolor{lightgray}
    \multicolumn{19}{c}{\textbf{Supervised Setting}} \\ \midrule
    \multirow{2}{*}{Methods}                            & \multirow{2}{*}{Pub.~~~}    & \multicolumn{1}{l|}{\multirow{2}{*}{Input Size}} & \multicolumn{4}{c|}{CAMO}                                                                                             & \multicolumn{4}{c|}{NC4K}                                                                                             & \multicolumn{4}{c|}{COD10K}                                                                                           & \multicolumn{4}{c}{CHAMELEON}                                                                                         \\ \cline{4-19}  
    & & & $F_{\beta}^w \uparrow$ & $\text{MAE}\downarrow$ & $S_{\alpha}\uparrow$ & $E_{\phi}\uparrow$ & $F_{\beta}^w \uparrow$ & $\text{MAE}\downarrow$ & $S_{\alpha}\uparrow$ & $E_{\phi}\uparrow$ & $F_{\beta}^w \uparrow$ & $\text{MAE}\downarrow$ & $S_{\alpha}\uparrow$ & $E_{\phi}\uparrow$ & $F_{\beta}^w \uparrow$ & $\text{MAE}\downarrow$ & $S_{\alpha}\uparrow$ & $E_{\phi}\uparrow$ \\ \midrule
    \multicolumn{1}{l|}{SINet \cite{SINet}}              & \multicolumn{1}{r|}{CVPR20}  & 352                                               & .606                        & .100                        & .751                        & .829                        & .723                        & .058                        & .808                        & .871                        & .551                        & .051                        & .771                        & .806                        & .740                        & .444                        & .869                        & .891                        \\
    \multicolumn{1}{l|}{LSR \cite{NC4K}}                 & \multicolumn{1}{r|}{CVPR21}  & 352                                               & .696                        & .105                        & .787                        & .854                        & .766                        & .048                        & .840                        & .907                        & .673                        & .037                        & .804                        & .892                        & .673                        & .037                        & .804                        & .892                        \\
    \multicolumn{1}{l|}{ZoomNet \cite{ZoomNet}}          & \multicolumn{1}{r|}{CVPR22}  & 384                                               & .752                        & .066                        & .820                        & .892                        & .784                        & .043                        & .853                        & .912                        & .729                        & .029                        & .838                        & .911                        & .845                        & {\color[HTML]{009901} .023}                        & .902                        & {\color[HTML]{009901} .958}                        \\
    \multicolumn{1}{l|}{SegMaR \cite{SegmMaR}}           & \multicolumn{1}{r|}{CVPR22}  & 352                                               & .724                        & .072                        & .805                        & .864                        & .781                        & .046                        & .841                        & .907                        & .686                        & .036                        & .813                        & .890                        & .828                        & .032                        & .888                        & .935                        \\
    \multicolumn{1}{l|}{SINet-V2 \cite{SINetV2}}         & \multicolumn{1}{r|}{TPAMI22} & 352                                               & .743                        & .070                        & .820                        & .895                        & .790                        & .060                        & .822                        & .885                        & .680                        & .037                        & .815                        & .906                        & .816                        & .030                        & .888                        & .942                        \\
    \multicolumn{1}{l|}{OSFormer \cite{OSFormer}}        & \multicolumn{1}{r|}{ECCV22}  & 384                                               & .767                        & .073                        & .799                        & .858                        & .790                        & .049                        & .832                        & .891                        & .701                        & .034                        & .811                        & .811                        & .836                        & .028                        & .891                        & .939                        \\
    \multicolumn{1}{l|}{HitNet \cite{HitNet}}            & \multicolumn{1}{r|}{AAAI23}  & 704                                               & .806                        & .056                        & .844                        & .904                        & -                           & -                           & -                           & -                           & {.804}                        & {\color[HTML]{3166FF} .023}  & .869                        & {\color[HTML]{009901} .936}  & -                           & -                           & -                           & -                           \\
    \multicolumn{1}{l|}{PENet \cite{PENet}}              & \multicolumn{1}{r|}{IJCAI23} & 352                                               & .771                        & .063                        & .828                        & .890                        & .795                        & .042                        & .855                        & .912                        & .723                        & .031                        & .831                        & .908                        & .851                        & {.024}  & .902                        & {\color[HTML]{3166FF} .960}  \\
    \multicolumn{1}{l|}{DaCOD \cite{DaCOD}}              & \multicolumn{1}{r|}{MM23}    & 448                                               & .796                        & {.051}                        & .855                        & .902                        & .814                        & .036                        & .874                        & .912                        & .729                        & .028                        & .840                        & .906                        & -                           & -                           & -                           & -                           \\
    \multicolumn{1}{l|}{FEDER \cite{FEDER}}              & \multicolumn{1}{r|}{CVPR23}  & 384                                               & .807                        & .066                        & .836                        & .897                        & .824                        & .042                        & .862                        & .913                        & .748                        & .029                        & .844                        & .911                        & {.856}                        &  .026  & {.903}                       & .947                        \\
    \multicolumn{1}{l|}{FSPNet \cite{FSPNet}}            & \multicolumn{1}{r|}{CVPR23}  & 384                                               & .799                        & {\color[HTML]{009901} .050} & .856                        & .899                        & .816                        & {.035} & .879                        & .915                        & .735                        & .026                        & .851                        & .895                        & -                           & -                           & -                           & -                           \\
    \multicolumn{1}{l|}{EVP-Segformer \cite{EVP,SegFormer}}                  & \multicolumn{1}{r|}{CVPR23}  & 352                                               & .777                        & .059                        & .846                        & .895                        & .818                        & .047                        & .855                        & .901                        & .742                        & .029                        & .843                        & .907                        & .795                        & .036                        & .871                        & .917                        \\
    \multicolumn{1}{l|}{ZoomNeXt \cite{zoomnext}}        & \multicolumn{1}{r|}{TPAMI24} & 384                                               & {\color[HTML]{009901} .857} & {\color[HTML]{3166FF} .041} & {\color[HTML]{009901} .889} & {\color[HTML]{CB0000} .945} & {\color[HTML]{009901} .863} & {\color[HTML]{CB0000} .028} & {\color[HTML]{009901} .903} & {\color[HTML]{3166FF} .951} & {\color[HTML]{3166FF} .827} & {\color[HTML]{CB0000} .018}  & {\color[HTML]{CB0000} .898} & {\color[HTML]{CB0000} .956}  & {\color[HTML]{3166FF} .885} & {\color[HTML]{CB0000} .018}  & {\color[HTML]{CB0000} .924} & {\color[HTML]{CB0000} .975}  \\ 
    \multicolumn{1}{l|}{SENet \cite{SENet}}        & \multicolumn{1}{r|}{TIP24} & 384                                               & .847                                               & {\color[HTML]{CB0000} .039}                   & .888                 & .932                                    & .843                   & {\color[HTML]{009901} .032}                   & .889                 & {\color[HTML]{009901} .933}               & .780                   & {\color[HTML]{009901} .024}                   & .865                 & .925               & .878                   & {\color[HTML]{3166FF} .019}                   & {\color[HTML]{3166FF} .918}                 & .957   \\ 
    \multicolumn{1}{l|}{COMPrompter \cite{COMPrompter}}        & \multicolumn{1}{r|}{SCIS24} & 1024                                               & {\color[HTML]{3166FF} .858}                                               & .044                   & .882                 & {\color[HTML]{3166FF} .942}                                    & {\color[HTML]{3166FF} .876}                   & {\color[HTML]{3166FF} .030}                   & {\color[HTML]{CB0000} .907}                 & {\color[HTML]{CB0000} .955}               & {\color[HTML]{009901} .821}                   & {\color[HTML]{3166FF} .023}                   & {\color[HTML]{009901} .889}                 & {\color[HTML]{3166FF} .949}               & {\color[HTML]{009901} .857}                   & .026                   & .906                 & { .955}    \\ 
    \multicolumn{1}{l|}{CDP \cite{CDP}}        & \multicolumn{1}{r|}{ESWA25} & 320                                               & .767                                               & .071                   & .798                 & .844                                    & .799                   & .051                   & .880                 & .902               & .709                   & .036                   & .791                 & .863               & .819                   & .033                   & .865                 & .923   \\ 
    \multicolumn{1}{l|}{EVP-EVA02-L \cite{EVP,EVA02}}        & \multicolumn{1}{c|}{-} & 352                                            & {.838} & {\color[HTML]{009901} .050} & {\color[HTML]{3166FF} .895} & {.924} & {.854} & .037 & {.902} & {\color[HTML]{009901}  .933} & .784 & .030  & {.878} & {.924}  &  .836 & .035  & .901 & .937  \\
    
    \midrule
    \rowcolor{tablecl} CAMF & \multicolumn{1}{c|}{-}       & 384                                               & {\color[HTML]{CB0000} .887} & {\color[HTML]{3166FF} .041}  & {\color[HTML]{CB0000} .901}  & {\color[HTML]{009901} .933}  & {\color[HTML]{CB0000} .897}  & {.033}  & {\color[HTML]{3166FF} .905}  & {\color[HTML]{009901} .933}  & {\color[HTML]{CB0000} .845}  & {.025}  & {\color[HTML]{3166FF} .890}  & .923  & {\color[HTML]{CB0000} .895}  & .028                        & {\color[HTML]{009901} .917}  & .944                        \\ \bottomrule
    \end{tabular}
    }
    \label{tab:results}
\end{table*}

In the context of supervised learning, we evaluated our framework against 14 other methods across four different datasets. As shown in Table~\ref{tab:results}, our framework consistently outperforms the alternatives, with particularly strong results on the CAMO and NC4K datasets, where it achieves the highest weighted F-measure scores. This demonstrates our model’s superior ability to confidently detect camouflaged objects in these challenging datasets.

On the COD10K and CHAMELEON dataset, our framework achieves the highest weighted F-measure scores of 0.845 and 0.895 respectively; however, it slightly trails other methods, such as ZoomNext~\cite{zoomnext} and HitNet~\cite{HitNet}, particularly in S-measure and E-measure. This performance discrepancy may be attributed to ZoomNext utilizing three different resolutions derived from the original 384-resolution image as raw inputs, and HitNet employs a high resolution of 704 to enhance segmentation detail.
Overall, these experimental results clearly demonstrate that our approach surpasses many other supervised learning models in tackling the COS task, delivering competitive performance on multiple datasets.

In Figure~\ref{fig:s_visual}, we present visual comparisons between our method and other COS approaches under supervised learning setting. The results indicate that our proposed method outperforms the alternative approaches across various levels of difficulty in mask prediction for the majority of samples.

\subsection{Ablation Study}
\label{sec:abl}

\begin{table*}[]
    \centering
    \caption{Ablation study on various aspects on the weighted F-measure ($F_{\beta}^w$). The rows highlighted in gray represent our method (default setting).}
    \setlength{\tabcolsep}{2.5pt}
    \resizebox{0.9\linewidth}{!}{
   \begin{tabular}{@{}ll|cccc|cccc@{}}
\toprule
\multicolumn{2}{c|}{\multirow{2}{*}{\centering \textbf{Method}}}                                                        & \multicolumn{4}{c|}{\textbf{Supervised}}                                                                                                            & \multicolumn{4}{c}{\textbf{Zero-Shot}}                                                                                                              \\ \cline{3-10} 
\multicolumn{2}{c|}{}                                                                                                   & \textbf{COD10K}                     & \textbf{CAMO}                      & \textbf{NC4K}                      & \textbf{CHAMELEON}                  & \textbf{COD10K}                     & \textbf{CAMO}                      & \textbf{NC4K}                      & \textbf{CHAMELEON}                  \\ \midrule
\multicolumn{1}{l|}{\multirow{2}{*}{\textbf{Caption Strategy}}} & M-LLM Caption                                         & .826                                & .872                               & .886                               & .877                                & .708                                & .720                               & .809                               & .712                                \\ 
\multicolumn{1}{l|}{}                                           & \cellcolor{tablecl}Codebook                           & \cellcolor{tablecl}\textbf{.845}  & \cellcolor{tablecl}\textbf{.887} & \cellcolor{tablecl}\textbf{.897} & \cellcolor{tablecl}\textbf{.895}  & \cellcolor{tablecl}\textbf{.717}  & \cellcolor{tablecl}\textbf{.729} & \cellcolor{tablecl}\textbf{.818} & \cellcolor{tablecl}\textbf{.720}  \\ \cline{1-10} 
\multicolumn{1}{l|}{\multirow{4}{*}{\textbf{MFA Component}}}    & Linear w/o multi-scale                                & .819                                & .867                               & .879                               & .877                                & .669                                & .664                               & .781                               & .670                                \\
\multicolumn{1}{l|}{}                                           & MLP Proj. \& Mixer w/o multi-scale                    & .816                                & .878                               & .873                               & .871                                & .698                                & \textbf{.730}                      & .797                               & .704                                \\
\multicolumn{1}{l|}{}                                           & Linear + multi-scale                                  & .845                                & .878                               & .896                               & \textbf{.895}                       & .696                                & .673                               & .800                               & .673                                \\
\multicolumn{1}{l|}{}                                           & \cellcolor{tablecl}MLP Proj. \& Mixer + multi-scale & \cellcolor{tablecl} \textbf{.845} & \cellcolor{tablecl}\textbf{.887} & \cellcolor{tablecl}\textbf{.897} & \cellcolor{tablecl} \textbf{.895} & \cellcolor{tablecl}\textbf{.717}  & \cellcolor{tablecl}.729          & \cellcolor{tablecl}\textbf{.818} & \cellcolor{tablecl}\textbf{.720}  \\ \cline{1-10}
\multicolumn{1}{l|}{\multirow{4}{*}{\textbf{Channels}}}         & 64 Channels (72.0K MFA Params)                        & .799                                & .841                               & .842                               & .837                                & .691                                & .678                               & .800                               & .692                                \\
\multicolumn{1}{l|}{}                                           & 128 Channels (0.204M MFA Params)                      & .819                                & .853                               & .860                               & .859                                & .714                                & .710                               & .818                               & .705                                \\
\multicolumn{1}{l|}{}                                           & \cellcolor{tablecl}256 Channels (0.664M MFA Params) & \cellcolor{tablecl} \textbf{.845} & \cellcolor{tablecl}\textbf{.887} & \cellcolor{tablecl}\textbf{.897} & \cellcolor{tablecl}\textbf{.895}  & \cellcolor{tablecl} \textbf{.717} & \cellcolor{tablecl}\textbf{.729} & \cellcolor{tablecl}\textbf{.818} & \cellcolor{tablecl}\textbf{.720}  \\
\multicolumn{1}{l|}{}                                           & 512 Channels (2.37M MFA Params)                       & .841                                & .882                               & .889                               & .893                                & .498                                & .517                               & .621                               & .536                                \\ \cline{1-10}
\multicolumn{1}{l|}{\multirow{2}{*}{\textbf{PEFT Module}}}      & LoRA                                                  & .824                                & .881                               & .878                               & .878                                & .692                                & .677                               & .798                               & .633                                \\
\multicolumn{1}{l|}{}                                           & \cellcolor{tablecl}Adapter                         & \cellcolor{tablecl}\textbf{.845}  & \cellcolor{tablecl}\textbf{.887} & \cellcolor{tablecl}\textbf{.897} & \cellcolor{tablecl}\textbf{.895}  & \cellcolor{tablecl}\textbf{.717}  & \cellcolor{tablecl}\textbf{.729} & \cellcolor{tablecl}\textbf{.818} & \cellcolor{tablecl} \textbf{.720} \\ \bottomrule
\end{tabular}
    }
    \label{tab:peft}
\end{table*}

\begin{figure}[t]
    \centering
    \includegraphics[width=0.9\linewidth]{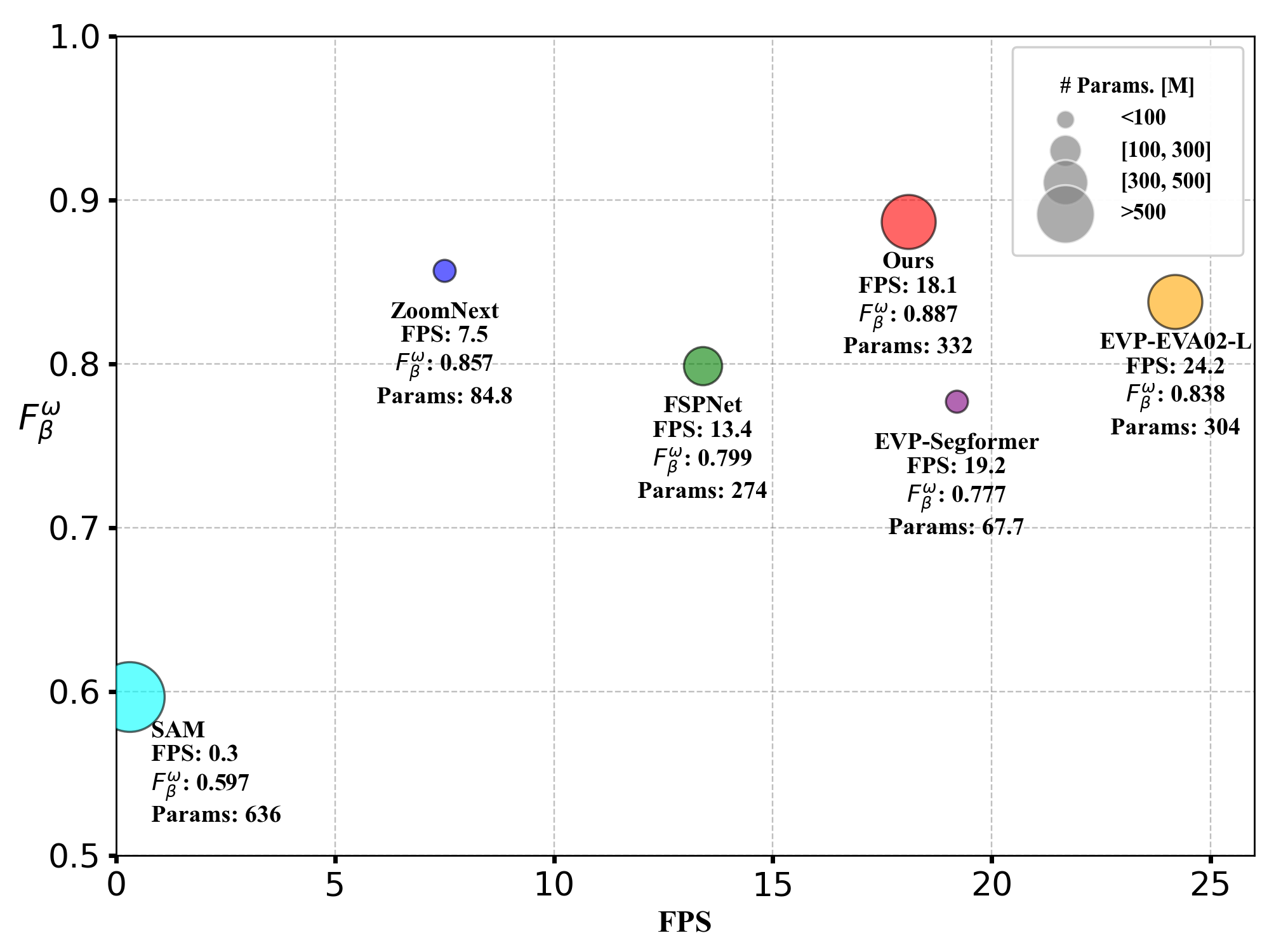}
    \caption{Comparison of models based on $F_{\beta}^w$ score, FPS, and number of parameters. Larger markers indicate models with more parameters. }
    \label{fig:fps}
\end{figure}

Our method employs a straightforward alignment strategy through the use of MFA. In order to assess the effectiveness of each component within MFA, we conducted ablation studies on COS datasets.
\noindent {\bf The Caption Strategy.}
In Table~\ref{tab:peft}, we compare two captioning strategies for the inference phase: captions generated from the M-LLM and those derived from the codebook, utilizing the weighted F-measure metric. Our findings indicate that the codebook consistently outperforms captions from the M-LLM across all datasets and application scenarios, demonstrating its overall superior performance.
\noindent {\bf The MFA Components.}
In Table~\ref{tab:peft}, we evaluated four different approaches for processing and analyzing image features: using a Linear Projector alone, utilizing an MLP Projector alone, and incorporating multi-scale information for image processing with both methods separately. The combined approach of Multi-scale and MLP Projector demonstrates superior performance.
%
%
This suggests that integrating multi-resolution inputs effectively improves model accuracy. Thus, we selected the MLP Proj. \& Mixer + multi-scale approach. 
%
%


\noindent {\bf The Channels of MFA Projector and Mixer.}
Furthermore, Table~\ref{tab:peft} compared the performance of four different channel parameters corresponding to feature projector and text mixer. After considering the trade-off between performance and the number of parameters, we set channels to 256 to achieve an optimal balance.

\noindent {\bf The PEFT Module.}
We evaluated two different Parameter-Efficient Fine-Tuning (PEFT) modules: LoRA \cite{LoRA} and Adapter \cite{AdaptFormer}. The results were tested across four COS datasets. In Table~\ref{tab:peft}, we present the performance of models using LoRA and Adapter in both supervised learning and zero-shot settings. Generally, Adapter outperforms LoRA. This may be due to the tuning of LoRA potentially obscuring local patterns learned during the MIM pre-training, as these local features might be influenced by shape and semantic features from salient objects. In contrast, Adapter does not face this issue, as the FFN in the transformer block acts as a key-value memories for attention \cite{kv_mem}. By using Adapter, we can effectively learn shape and semantic features without compromising the local patterns established during the MIM phase, enabling better extraction of useful features for zero-shot COS.

\begin{table}[]

    \centering

    \caption{Ablation study examining the influence of different prompts during training on the weighted F-measure ($F^w_{\beta}$) for CAMO and COD10K datasets. In zero-shot setting, we use SOS task prompt. In supervised leanrning setting, we use COS task prompt. The results demonstrate how prompt selection affects model performance.}
    \setlength{\tabcolsep}{5pt}

    \resizebox{0.9\linewidth}{!}{
    \begin{tabular}{@{}llll@{}}
\toprule
\multicolumn{1}{l|}{ID} & \multicolumn{1}{l|}{Prompt}                             & CAMO & COD10K \\ \midrule
\rowcolor{lightgray}\multicolumn{4}{c}{SOS}                                                                           \\ \midrule
\multicolumn{1}{l|}{1}  & \multicolumn{1}{l|}{"The  object."}                     & .697 & .693   \\
\multicolumn{1}{l|}{2}  & \multicolumn{1}{l|}{"The salient object."}              & .693 & .694   \\
\multicolumn{1}{l|}{3}  & \multicolumn{1}{l|}{"Which object is salient?"}         & .697 & .692   \\
\multicolumn{1}{l|}{4}  & \multicolumn{1}{l|}{"Describe the image."}              & \textbf{.729} & \textbf{.717}   \\
\multicolumn{1}{l|}{5}  & \multicolumn{1}{l|}{"Describe the object."}             & .716 & .710   \\
\multicolumn{1}{l|}{6}  & \multicolumn{1}{l|}{"Describe the salient object."}     & .717 & .710   \\ \midrule
\rowcolor{lightgray}\multicolumn{4}{c}{COS}                                                                           \\ \midrule
\multicolumn{1}{l|}{7}  & \multicolumn{1}{l|}{"The camouflaged object."}          & .866 & .823   \\
\multicolumn{1}{l|}{8}  & \multicolumn{1}{l|}{"The camouflaged animal"}           & .879 & .838   \\
\multicolumn{1}{l|}{9}  & \multicolumn{1}{l|}{"Which animal is camouflaged?"}     & .879 & .840   \\
\multicolumn{1}{l|}{10} & \multicolumn{1}{l|}{"Describe the image."}              & .886 & .842   \\
\multicolumn{1}{l|}{11} & \multicolumn{1}{l|}{"Describe the camouflaged animal."} & \textbf{.887} & \textbf{.845}   \\ \bottomrule
\end{tabular}
    }
    \label{tab:prompt}
\end{table}

\noindent {\bf The Task Prompts.} In Table~\ref{tab:prompt}, we present the impact of various prompts on model performance. In the zero-shot setting, we used the SOS task prompt and observed that the instruction "Describe the image." yielded the best results. This can be attributed to the fact that describing the entire image allows the codebook to capture a more comprehensive and global representation. Conversely, limiting the description to only salient objects may constrain the semantic space, reducing the model's ability to generalize. In the supervised learning setting, we employed the COS task prompt, where "Describe the camouflaged animal." led to the highest performance. 

Interestingly, replacing "camouflaged animal" with "camouflaged object" introduced ambiguity between salient and camouflaged objects in the M-LLM's output, which negatively affected the model's accuracy. By specifying camouflaged objects and providing detailed descriptions, we enable the codebook to learn more effective representations tailored to these specific objects, resulting in improved model performance.

\subsection{Analysis of Running Times (FPS)}
Using the learned query during inference simplifies our pipeline, enabling it to achieve frame rates comparable to current state-of-the-art models. We evaluated our method on 100 images using a NVIDIA RTX 4060Ti GPU paired with an i7-13790F CPU. The results are presented in Figure~\ref{fig:fps}. The size of the markers in the plot reflects the number of parameters in each model, with larger markers representing models with more parameters. The plot highlights the trade-offs between model size, speed, and performance. For the performance, we report the \( F^{\omega}_{\beta} \) score for each method on the CAMO dataset.

\textbf{EVP-EVA02-L} achieves the highest FPS at 24.2, with an \( F^{\omega}_{\beta} \) score of 0.838 and 304 million parameters. \textbf{EVP-Segformer} and \textbf{FSPNet} offer moderate FPS (19.2 and 13.4) and \( F^{\omega}_{\beta} \) scores (0.777 and 0.799), with parameter counts of 67.7M and 274M, respectively. \textbf{ZoomNext} provides a balance with 7.5 FPS and a high \( F^{\omega}_{\beta} \) score of 0.857, using 84.8 million parameters. \textbf{SAM}, with the largest parameter count (636M), operates at only 0.3 FPS, achieving an \( F^{\omega}_{\beta} \) score of 0.597. 
Finally, \textbf{Our method} delivers an impressive \( F^{\omega}_{\beta} \) score of 0.887 at 18.1 FPS, with 332M parameters. As illustrated, our approach strikes the best balance between performance, FPS, and parameter efficiency, offering competitive speed and accuracy with an affordable parameter size.





\begin{table}[t]

    \centering

    \caption{Additional experiment on polyp segmentation.}
    \setlength{\tabcolsep}{5pt}

    \resizebox{0.98\linewidth}{!}{
    \begin{tabular}{@{}lcccccccc@{}}
\toprule
\multicolumn{1}{l|}{\multirow{2}{*}{Methods}} & \multicolumn{4}{c}{Kvasir}                                                                                       & \multicolumn{4}{c}{ETIS}                                                                    \\ \cline{2-9} 
\multicolumn{1}{l|}{}                         & $F_{\beta}^w \uparrow$ & $\text{MAE}\downarrow$ & $S_{\alpha}\uparrow$ & \multicolumn{1}{c|}{$E_{\phi}\uparrow$} & $F_{\beta}^w \uparrow$ & $\text{MAE}\downarrow$ & $S_{\alpha}\uparrow$ & $E_{\phi}\uparrow$ \\ \midrule
\rowcolor{lightgray}\multicolumn{9}{c}{Zero-shot}                                                                                                                                                                                                                                  \\ \midrule
\multicolumn{1}{l|}{CLIP Surgery+SAM \cite{SAM,clip_surgery}}          & -                      & -                      & \textbf{-}           & \multicolumn{1}{c|}{-}                  & .047                   & .537                   & .272                 & .296               \\
\multicolumn{1}{l|}{GenSAM \cite{gensam}}                   & -                      & -                      & -                    & \multicolumn{1}{c|}{-}                  & .090                   & .205                   & .430                 & .554               \\
\multicolumn{1}{l|}{CAMF (ZS)}                     & .426                   & .094                   & .633                 & \multicolumn{1}{c|}{.575}               & \textbf{.330}          & \textbf{.087}          & \textbf{.620}        & \textbf{.570}      \\ \midrule
\rowcolor{lightgray}\multicolumn{9}{c}{Supervised}                                                                                                                                                                                                                                 \\ \midrule
\multicolumn{1}{l|}{UNet \cite{unet}}                     & .794                   & .055                   & .858                 & \multicolumn{1}{c|}{.893}               & .366                   & .036                   & .843                 & .876               \\
\multicolumn{1}{l|}{PraNet \cite{fan2020pranet}}                   & .885                   & .030                   & .915                 & \multicolumn{1}{c|}{.948}               & .600                   & .031                   & .794                 & .841               \\
\multicolumn{1}{l|}{CAMF (S)}                     & \textbf{.977}          & \textbf{.008}          & \textbf{.973}        & \multicolumn{1}{c|}{\textbf{.982}}      & \textbf{.760}          & \textbf{.018}          & \textbf{.882}        & \textbf{.899}      \\ \bottomrule
\end{tabular}
    }
    \label{tab:polyp}
\end{table}

\section{Additional Experiments on Diverse Applications}

{To validate the generalizability of our framework, we conducted comprehensive experiments across multiple domains: (1) video camouflaged object segmentation \cite{moca,cad2016}, (2) image-based camouflaged object segmentation (using the newly released OVCAMO dataset) \cite{OVCOS}, (3) medical domain (polyp segmentation) \cite{jha2020kvasir}, and (4) underwater scene \cite{li2020mas3k,fu2023masnet}.

\noindent{\bf Extended Experiments on Camouflaged Objects.}
we have conducted additional experiments on these two datasets, MoCA-Mask \cite{moca} and OVCamo \cite{OVCOS}. As shown in Table~\ref{tab:moca_ovcamo}, our method achieves best performance under the zero-shot setting on both datasets. For OVCamo (originally designed for open-vocabulary tasks), we evaluate only its mask annotations under standard binary segmentation settings to ensure fair comparison. Our analysis uses currently available benchmark data, excluding the CAD~\cite{cad2016} subset which is no longer publicly accessible. This approach guarantees reproducible results and equitable method comparison using this version of OV-CAMO.}

\noindent{\bf Polyp Segmentation.}
In the zero-shot setting, we aimed to transfer knowledge from camouflaged object segmentation to polyp segmentation, utilizing our pre-trained COS model for direct inference on polyp segmentation tasks. For the supervised learning scenario, we trained the model using the Kvasir dataset \cite{jha2020kvasir} and evaluated its performance on the Kvasir and ETIS test sets \cite{etis}, which are widely used for benchmarking. The Kvasir dataset typically serves as the training set, while the ETIS test set is part of the standard evaluation suite. The results, as presented in Table \ref{tab:polyp}, demonstrate that our model achieves state-of-the-art performance in zero-shot polyp segmentation by effectively transferring knowledge from camouflaged object segmentation. Furthermore, our method achieved the highest weighted F-measure scores of 97.7\% on the Kvasir test set and 76.0\% on the ETIS test set.

{\noindent{\bf Underwater Scene.}
To evaluate cross-scene generalization beyond natural terrestrial images in the SOS dataset, we extend validation to underwater segmentation using MAS3K~\cite{li2020mas3k} and RMAS~\cite{fu2023masnet} datasets. For the supervised learning, we follow the training setting in \cite{mas_sam}. In evaluating zero-shot performance on underwater scenes, we directly employ our DUTS-trained model. As shown in Table~\ref{tab:underwater}, our method achieves competitive zero-shot performance, even outperforming supervised baselines like MASNet on MAS3K. Under supervised learning, our method delivers the best performance in both $F_{\beta}^w$ and $S_{\alpha}$ metrics. Specifically, we achieve 86.5\% $F_{\beta}^w$ and 89.7\% $S_{\alpha}$ on MAS3K, and 90.1\% $F_{\beta}^w$ with 86.8\% $S_{\alpha}$ on RMAS.}

\begin{table*}[t]
\centering
\caption{Comparison of our method with other zero-shot methods (\textbf{bold} indicates best performance under zero-shot setting), with additional supervised results of our method for reference (* denotes continual training on MoCA-MASK for 2 epochs). }
\resizebox{.9\linewidth}{!}{
    \begin{tabular}{@{}l|c|cccc|cccc@{}}
\toprule
\multirow{2}{*}{Method} & \multirow{2}{*}{\begin{tabular}[c]{@{}c@{}}Segmentor \\ Training Data\end{tabular}} & \multicolumn{4}{c|}{MoCA-MASK~\cite{moca}}                                                                                                       & \multicolumn{4}{c}{OVCamo~\cite{OVCOS}}                                                                                                          \\ \cmidrule(l){3-10} 
                        &                                                                                     & $F_{\beta}^w \uparrow$ & $\text{MAE}\downarrow$ & \multicolumn{1}{l}{$S_{\alpha}\uparrow$} & \multicolumn{1}{l|}{$E_{\phi}\uparrow$} & $F_{\beta}^w \uparrow$ & $\text{MAE}\downarrow$ & \multicolumn{1}{l}{$S_{\alpha}\uparrow$} & \multicolumn{1}{l}{$E_{\phi}\uparrow$} \\ \midrule
GenSAM~\cite{gensam}                  & SA-1B                                                                               & .141                   & .067                   & .523                                     & .243                                       & -   & -   & -                    & -                 \\
MMCPF~\cite{MMCPF}                  & SA-1B                                                                               & .196                   & .031                   & .569                                     & .451                                    & -   & -   & -                     & -                  \\ \midrule
VST~\cite{VST}                     & DUTS                                                                                & .099                   & .046                   & .565                                     & .488                                    & .498                   & .070                   & .700                                     & .756                                   \\
SINet-V2~\cite{SINetV2}                & DUTS                                                                                & .097                   & .058                   & .595                                     & .411                                    & .371                   & .080                   & .631                                     & .606                                   \\
EVP-Segformer~\cite{EVP,SegFormer}           & DUTS                                                                                & .126                   & .031                   & .581                                     & .486                                    & .610                   & .047                   & .751                                     & .760                                   \\
EVP-EVA02-L~\cite{EVP,EVA02}             & DUTS                                                                                & .103                   & .088                   & .536                                     & .463                                    & .591                   & .068                   & .755                                     & .783                                   \\
CAMF (ZS)             & DUTS                                                                                & \textbf{.263}          & \textbf{.023}          & \textbf{.671}                            & \textbf{.671}                           & \textbf{.664}          & \textbf{.050}          & \textbf{.790}                            & \textbf{.801}                          \\ \midrule
CAMF (S)             & COD10K + CAMO                                                                       & .510                   & .007                   & .774                                     & .774                                    & -                      & -                      & -                                        & -                                      \\
CAMF* (S)            & COD10K + CAMO + MoCA-MASK                                                           & .575                   & .008                   & .773                                     & .776                                    & -                      & -                      & -                                        & -                                      \\ \bottomrule
\end{tabular}
}
    \label{tab:moca_ovcamo}
\end{table*}

\begin{table}[t]

    \centering

    \caption{Results on underwater scene segmentation, evaluating our method under supervised (S) and zero-shot (ZS) training paradigms.}
    \setlength{\tabcolsep}{5pt}

    \resizebox{0.98\linewidth}{!}{
    \begin{tabular}{@{}l|cccc|cccc@{}}
\toprule
\multicolumn{1}{c|}{\multirow{2}{*}{Methods}} & \multicolumn{4}{c|}{MAS3K~\cite{li2020mas3k}}                                                                  & \multicolumn{4}{c}{RMAS~\cite{fu2023masnet}}                                                                    \\ \cmidrule(l){2-9} 
\multicolumn{1}{c|}{}                         & $F_{\beta}^w \uparrow$ & $\text{MAE}\downarrow$ & $S_{\alpha}\uparrow$ & $E_{\phi}\uparrow$ & $F_{\beta}^w \uparrow$ & $\text{MAE}\downarrow$ & $S_{\alpha}\uparrow$ & $E_{\phi}\uparrow$ \\ \midrule
ZoomNet~\cite{ZoomNet}                                       & .780                   & .032                   & .862                 & .898               & .795                   & .022                   & .855                 & .915               \\
MASNet~\cite{fu2023masnet}                                        & .788                   & .032                   & .864                 & .906               & .801                   & .024                   & .862                 & .920               \\
H2Former~\cite{he2023h2former}                                      & .810                   & .028                   & .865                 & .925               & .799                   & .023                   & .844                 & .931               \\
SAM~\cite{SAM,mas_sam}                                           & .656                   & .059                   & .763                 & .807               & .534                   & .053                   & .697                 & .790               \\
Dual-SAM~\cite{dual_sam}                                      & .838                   & \textbf{.023}          & .884                 & .933               & .812                   & .022                   & .860                 & .944               \\
MAS-SAM~\cite{mas_sam}                                       & .840                   & .025                   & .887                 & \textbf{.938}      & .819                   & \textbf{.021}          & .865                 & \textbf{.948}      \\
CAMF (S)                                         & \textbf{.865}          & .026                   & \textbf{.897}        & .906               & \textbf{.901}          & .027                   & \textbf{.868}        & .884               \\
CAMF (ZS)                                     & .803                   & .038                   & .867                 & .881               & .618                   & .067                   & .716                 & .712               \\ \bottomrule
\end{tabular}
    }
    \label{tab:underwater}
\end{table}

\section{Conclusion}

In this paper, we tackled the challenging problem of zero-shot Camouflaged Object Segmentation (ZSCOS) without the need for camouflaged object annotations. We proposed a novel framework that integrates the strengths of both vision and language models, leveraging a pre-trained Masked Image Modeling (MIM) encoder and Multimodal Large Language Models (M-LLM) to effectively align visual and semantic features. By incorporating Salient Object Segmentation (SOS) datasets and fine-grained alignment strategies, our approach bridges the gap between COS and SOS, enabling effective zero-shot transfer. Our key contributions also include the introduction of a codebook that substitutes the need for M-LLM during inference, significantly reducing computational overhead while maintaining high performance. Through extensive experiments, our method demonstrated state-of-the-art results in zero-shot COS, achieving impressive \( F^{\omega}_{\beta} \) scores on the CAMO and COD10K datasets, and competitive results in other domains such as polyp segmentation and underwater scene segmentation. Furthermore, we showed that our approach performs competitively in both zero-shot and supervised settings, highlighting its broad applicability across various segmentation tasks.

Beyond numerical advancements, our work paves the way for new possibilities in zero-shot learning for complex segmentation tasks, providing an efficient, scalable, and high-performance solution adaptable to diverse applications. The flexibility of our framework enables it to be applied across a broad spectrum of domains, from medical imaging to environmental monitoring, where annotated data is scarce or unavailable. Future research could focus on extending this approach to other challenging dense prediction tasks, utilizing larger datasets to provide richer semantic information, refining the codebook mechanism, and further optimizing the trade-off between computational efficiency and segmentation accuracy.

\section*{Acknowledgment}
This work was supported by  the Key Program for International Cooperation of Ministry of Science and Technology of China (No.2024YFE0100700) and the Key Project of the Natural Science Foundation of Sichuan Province (No.2025ZNSFSC0002).

\bibliographystyle{IEEEtran}

\begin{thebibliography}{100}
\providecommand{\url}[1]{#1}
\csname url@samestyle\endcsname
\providecommand{\newblock}{\relax}
\providecommand{\bibinfo}[2]{#2}
\providecommand{\BIBentrySTDinterwordspacing}{\spaceskip=0pt\relax}
\providecommand{\BIBentryALTinterwordstretchfactor}{4}
\providecommand{\BIBentryALTinterwordspacing}{\spaceskip=\fontdimen2\font plus
\BIBentryALTinterwordstretchfactor\fontdimen3\font minus \fontdimen4\font\relax}
\providecommand{\BIBforeignlanguage}[2]{{%
\expandafter\ifx\csname l@#1\endcsname\relax
\typeout{** WARNING: IEEEtran.bst: No hyphenation pattern has been}%
\typeout{** loaded for the language `#1'. Using the pattern for}%
\typeout{** the default language instead.}%
\else
\language=\csname l@#1\endcsname
\fi
#2}}
\providecommand{\BIBdecl}{\relax}
\BIBdecl

\bibitem{SENet}
C.~Hao, Z.~Yu, X.~Liu, J.~Xu, H.~Yue, and J.~Yang, ``A simple yet effective network based on vision transformer for camouflaged object and salient object detection,'' \emph{IEEE Transactions on Image Processing}, 2025.

\bibitem{CDP}
Y.~Liu, C.~Li, X.~Dong, L.~Li, D.~Zhang, S.~Xu, and J.~Han, ``Seamless detection: Unifying salient object detection and camouflaged object detection,'' \emph{Expert Systems with Applications}, vol. 274, p. 126912, 2025.

\bibitem{COMPrompter}
X.~Zhang, Z.~Yu, L.~Zhao, D.-P. Fan, and G.~Xiao, ``Comprompter: reconceptualized segment anything model with multiprompt network for camouflaged object detection,'' \emph{Science China Information Sciences}, vol.~68, no.~1, p. 112104, 2025.

\bibitem{yang2023silt}
H.~Yang, T.~Wang, X.~Hu, and C.-W. Fu, ``Silt: Shadow-aware iterative label tuning for learning to detect shadows from noisy labels,'' in \emph{Proceedings of the IEEE/CVF international conference on computer vision}, 2023, pp. 12\,687--12\,698.

\bibitem{fdrnet}
L.~Zhu, K.~Xu, Z.~Ke, and R.~W. Lau, ``Mitigating intensity bias in shadow detection via feature decomposition and reweighting,'' in \emph{Proceedings of the IEEE/CVF international conference on computer vision}, 2021, pp. 4702--4711.

\bibitem{tang2024divide}
H.~Tang, Z.~Li, D.~Zhang, S.~He, and J.~Tang, ``Divide-and-conquer: Confluent triple-flow network for rgb-t salient object detection,'' \emph{IEEE Transactions on Pattern Analysis and Machine Intelligence}, 2024.

\bibitem{sun2022singular}
Y.~Sun, Q.~Chen, X.~He, J.~Wang, H.~Feng, J.~Han, E.~Ding, J.~Cheng, Z.~Li, and J.~Wang, ``Singular value fine-tuning: Few-shot segmentation requires few-parameters fine-tuning,'' \emph{Advances in neural information processing systems}, vol.~35, pp. 37\,484--37\,496, 2022.

\bibitem{peng2019few}
Z.~Peng, Z.~Li, J.~Zhang, Y.~Li, G.-J. Qi, and J.~Tang, ``Few-shot image recognition with knowledge transfer,'' in \emph{Proceedings of the IEEE/CVF international conference on computer vision}, 2019, pp. 441--449.

\bibitem{mtmt_net}
Z.~Chen, L.~Zhu, L.~Wan, S.~Wang, W.~Feng, and P.-A. Heng, ``A multi-task mean teacher for semi-supervised shadow detection,'' in \emph{Proceedings of the IEEE/CVF Conference on computer vision and pattern recognition}, 2020, pp. 5611--5620.

\bibitem{mas_sam}
\BIBentryALTinterwordspacing
T.~Yan, Z.~Wan, X.~Deng, P.~Zhang, Y.~Liu, and H.~Lu, ``Mas-sam: segment any marine animal with aggregated features,'' in \emph{Proceedings of the Thirty-Third International Joint Conference on Artificial Intelligence}, ser. IJCAI '24, 2024. [Online]. Available: \url{https://doi.org/10.24963/ijcai.2024/761}
\BIBentrySTDinterwordspacing

\bibitem{dual_sam}
P.~Zhang, T.~Yan, Y.~Liu, and H.~Lu, ``Fantastic animals and where to find them: Segment any marine animal with dual sam,'' in \emph{Proceedings of the IEEE/CVF Conference on Computer Vision and Pattern Recognition}, 2024, pp. 2578--2587.

\bibitem{he2023h2former}
A.~He, K.~Wang, T.~Li, C.~Du, S.~Xia, and H.~Fu, ``H2former: An efficient hierarchical hybrid transformer for medical image segmentation,'' \emph{IEEE Transactions on Medical Imaging}, vol.~42, no.~9, pp. 2763--2775, 2023.

\bibitem{cad2016}
P.~Bideau and E.~Learned-Miller, ``It’s moving! a probabilistic model for causal motion segmentation in moving camera videos,'' in \emph{Computer Vision--ECCV 2016: 14th European Conference, Amsterdam, The Netherlands, October 11-14, 2016, Proceedings, Part VIII 14}.\hskip 1em plus 0.5em minus 0.4em\relax Springer, 2016, pp. 433--449.

\bibitem{istd}
J.~Wang, X.~Li, and J.~Yang, ``Stacked conditional generative adversarial networks for jointly learning shadow detection and shadow removal,'' in \emph{Proceedings of the IEEE conference on computer vision and pattern recognition}, 2018, pp. 1788--1797.

\bibitem{sbu}
T.~F.~Y. Vicente, L.~Hou, C.-P. Yu, M.~Hoai, and D.~Samaras, ``Large-scale training of shadow detectors with noisily-annotated shadow examples,'' in \emph{Computer Vision--ECCV 2016: 14th European Conference, Amsterdam, The Netherlands, October 11-14, 2016, Proceedings, Part VI 14}.\hskip 1em plus 0.5em minus 0.4em\relax Springer, 2016, pp. 816--832.

\bibitem{fu2023masnet}
Z.~Fu, R.~Chen, Y.~Huang, E.~Cheng, X.~Ding, and K.-K. Ma, ``Masnet: A robust deep marine animal segmentation network,'' \emph{IEEE Journal of Oceanic Engineering}, 2023.

\bibitem{li2020mas3k}
L.~Li, E.~Rigall, J.~Dong, and G.~Chen, ``Mas3k: An open dataset for marine animal segmentation,'' in \emph{International Symposium on Benchmarking, Measuring and Optimization}.\hskip 1em plus 0.5em minus 0.4em\relax Springer, 2020, pp. 194--212.

\bibitem{moca}
X.~Cheng, H.~Xiong, D.-P. Fan, Y.~Zhong, M.~Harandi, T.~Drummond, and Z.~Ge, ``Implicit motion handling for video camouflaged object detection,'' in \emph{CVPR}, 2022.

\bibitem{sam2zscod}
L.~Tang and B.~Li, ``Evaluating sam2's role in camouflaged object detection: From sam to sam2,'' \emph{arXiv preprint arXiv:2407.21596}, 2024.

\bibitem{MonaAdapter2}
\BIBentryALTinterwordspacing
D.~Yin, L.~Hu, B.~Li, Y.~Zhang, and X.~Yang, ``5
\BIBentrySTDinterwordspacing

\bibitem{ravi2024sam2}
\BIBentryALTinterwordspacing
N.~Ravi, V.~Gabeur, Y.-T. Hu, R.~Hu, C.~Ryali, T.~Ma, H.~Khedr, R.~R{\"a}dle, C.~Rolland, L.~Gustafson, E.~Mintun, J.~Pan, K.~V. Alwala, N.~Carion, C.-Y. Wu, R.~Girshick, P.~Doll{\'a}r, and C.~Feichtenhofer, ``Sam 2: Segment anything in images and videos,'' \emph{arXiv preprint arXiv:2408.00714}, 2024. [Online]. Available: \url{https://arxiv.org/abs/2408.00714}
\BIBentrySTDinterwordspacing

\bibitem{fan2020pranet}
D.-P. Fan, G.-P. Ji, T.~Zhou, G.~Chen, H.~Fu, J.~Shen, and L.~Shao, ``Pranet: Parallel reverse attention network for polyp segmentation,'' in \emph{International conference on medical image computing and computer-assisted intervention}.\hskip 1em plus 0.5em minus 0.4em\relax Springer, 2020, pp. 263--273.

\bibitem{unet}
O.~Ronneberger, P.~Fischer, and T.~Brox, ``U-net: Convolutional networks for biomedical image segmentation,'' in \emph{Medical image computing and computer-assisted intervention--MICCAI 2015: 18th international conference, Munich, Germany, October 5-9, 2015, proceedings, part III 18}.\hskip 1em plus 0.5em minus 0.4em\relax Springer, 2015, pp. 234--241.

\bibitem{clip_surgery}
Y.~Li, H.~Wang, Y.~Duan, and X.~Li, ``Clip surgery for better explainability with enhancement in open-vocabulary tasks,'' \emph{arXiv preprint arXiv:2304.05653}, 2023.

\bibitem{etis}
J.~Silva, A.~Histace, O.~Romain, X.~Dray, and B.~Granado, ``Toward embedded detection of polyps in wce images for early diagnosis of colorectal cancer,'' \emph{International journal of computer assisted radiology and surgery}, vol.~9, pp. 283--293, 2014.

\bibitem{cvc612}
J.~Bernal, F.~J. S{\'a}nchez, G.~Fern{\'a}ndez-Esparrach, D.~Gil, C.~Rodr{\'\i}guez, and F.~Vilari{\~n}o, ``Wm-dova maps for accurate polyp highlighting in colonoscopy: Validation vs. saliency maps from physicians,'' \emph{Computerized medical imaging and graphics}, vol.~43, pp. 99--111, 2015.

\bibitem{jha2020kvasir}
D.~Jha, P.~H. Smedsrud, M.~A. Riegler, P.~Halvorsen, T.~De~Lange, D.~Johansen, and H.~D. Johansen, ``Kvasir-seg: A segmented polyp dataset,'' in \emph{MultiMedia modeling: 26th international conference, MMM 2020, Daejeon, South Korea, January 5--8, 2020, proceedings, part II 26}.\hskip 1em plus 0.5em minus 0.4em\relax Springer, 2020, pp. 451--462.

\bibitem{gensam}
J.~Hu, J.~Lin, S.~Gong, and W.~Cai, ``Relax image-specific prompt requirement in sam: A single generic prompt for segmenting camouflaged objects,'' in \emph{Proceedings of the AAAI Conference on Artificial Intelligence}, vol.~38, no.~11, 2024, pp. 12\,511--12\,518.

\bibitem{MMCPF}
L.~Tang, P.-T. Jiang, Z.~Shen, H.~Zhang, J.~Chen, and B.~Li, ``Chain of visual perception: Harnessing multimodal large language models for zero-shot camouflaged object detection,'' in \emph{ACM Multimedia 2024}, 2024.

\bibitem{kv_mem}
M.~Geva, R.~Schuster, J.~Berant, and O.~Levy, ``Transformer feed-forward layers are key-value memories,'' \emph{arXiv preprint arXiv:2012.14913}, 2020.

\bibitem{dao2022flashattention}
T.~Dao, D.~Fu, S.~Ermon, A.~Rudra, and C.~R{\'e}, ``Flashattention: Fast and memory-efficient exact attention with io-awareness,'' \emph{Advances in Neural Information Processing Systems}, vol.~35, pp. 16\,344--16\,359, 2022.

\bibitem{relu}
X.~Glorot, A.~Bordes, and Y.~Bengio, ``Deep sparse rectifier neural networks,'' in \emph{Proceedings of the fourteenth international conference on artificial intelligence and statistics}.\hskip 1em plus 0.5em minus 0.4em\relax JMLR Workshop and Conference Proceedings, 2011, pp. 315--323.

\bibitem{DaCOD}
Q.~Wang, J.~Yang, X.~Yu, F.~Wang, P.~Chen, and F.~Zheng, ``Depth-aided camouflaged object detection,'' in \emph{ACMMM}, 2023, pp. 3297--3306.

\bibitem{xie2023revealing_mim}
Z.~Xie, Z.~Geng, J.~Hu, Z.~Zhang, H.~Hu, and Y.~Cao, ``Revealing the dark secrets of masked image modeling,'' in \emph{Proceedings of the IEEE/CVF conference on computer vision and pattern recognition}, 2023, pp. 14\,475--14\,485.

\bibitem{mae}
K.~He, X.~Chen, S.~Xie, Y.~Li, P.~Doll{\'a}r, and R.~Girshick, ``Masked autoencoders are scalable vision learners,'' in \emph{Proceedings of the IEEE/CVF conference on computer vision and pattern recognition}, 2022, pp. 16\,000--16\,009.

\bibitem{plain_detr}
Y.~Lin, Y.~Yuan, Z.~Zhang, C.~Li, N.~Zheng, and H.~Hu, ``Detr does not need multi-scale or locality design,'' in \emph{Proceedings of the IEEE/CVF International Conference on Computer Vision}, 2023, pp. 6545--6554.

\bibitem{SCWSSOD}
S.~Yu, B.~Zhang, J.~Xiao, and E.~G. Lim, ``Structure-consistent weakly supervised salient object detection with local saliency coherence,'' in \emph{AAAI}, vol.~35, no.~4, 2021, pp. 3234--3242.

\bibitem{TEL}
Z.~Liang, T.~Wang, X.~Zhang, J.~Sun, and J.~Shen, ``Tree energy loss: Towards sparsely annotated semantic segmentation,'' in \emph{CVPR}, 2022, pp. 16\,907--16\,916.

\bibitem{boundary2022}
H.~Zhu, P.~Li, H.~Xie, X.~Yan, D.~Liang, D.~Chen, M.~Wei, and J.~Qin, ``I can find you! boundary-guided separated attention network for camouflaged object detection,'' in \emph{Proceedings of the AAAI conference on artificial intelligence}, vol.~36, no.~3, 2022, pp. 3608--3616.

\bibitem{texture2023}
G.-P. Ji, D.-P. Fan, Y.-C. Chou, D.~Dai, A.~Liniger, and L.~Van~Gool, ``Deep gradient learning for efficient camouflaged object detection,'' \emph{Machine Intelligence Research}, vol.~20, no.~1, pp. 92--108, 2023.

\bibitem{texture2021}
J.~Zhu, X.~Zhang, S.~Zhang, and J.~Liu, ``Inferring camouflaged objects by texture-aware interactive guidance network,'' in \emph{Proceedings of the AAAI conference on artificial intelligence}, vol.~35, no.~4, 2021, pp. 3599--3607.

\bibitem{glass_detection2020}
H.~Mei, X.~Yang, Y.~Wang, Y.~Liu, S.~He, Q.~Zhang, X.~Wei, and R.~W. Lau, ``Don't hit me! glass detection in real-world scenes,'' in \emph{Proceedings of the IEEE/CVF Conference on Computer Vision and Pattern Recognition}, 2020, pp. 3687--3696.

\bibitem{Transparent2020}
E.~Xie, W.~Wang, W.~Wang, M.~Ding, C.~Shen, and P.~Luo, ``Segmenting transparent objects in the wild,'' in \emph{Computer Vision--ECCV 2020: 16th European Conference, Glasgow, UK, August 23--28, 2020, Proceedings, Part XIII 16}.\hskip 1em plus 0.5em minus 0.4em\relax Springer, 2020, pp. 696--711.

\bibitem{CI_CD_1982}
L.~W. Barsalou, ``Context-independent and context-dependent information in concepts,'' \emph{Memory \& cognition}, vol.~10, no.~1, pp. 82--93, 1982.

\bibitem{CI_CD_2018}
C.~Martial, D.~Stawarczyk, and A.~D'Argembeau, ``Neural correlates of context-independent and context-dependent self-knowledge,'' \emph{Brain and Cognition}, vol. 125, pp. 23--31, 2018.

\bibitem{CI_CD_2005}
T.~Lachmann and C.~Van~Leeuwen, ``Individual pattern representations are context independent, but their collectiverepresentation is context dependent,'' \emph{The Quarterly Journal of Experimental Psychology Section A}, vol.~58, no.~7, pp. 1265--1294, 2005.

\bibitem{zhao2024spider}
X.~Zhao, Y.~Pang, W.~Ji, B.~Sheng, J.~Zuo, L.~Zhang, and H.~Lu, ``Spider: A unified framework for context-dependent concept understanding,'' \emph{arXiv preprint arXiv:2405.01002}, 2024.

\bibitem{adamw}
I.~Loshchilov and F.~Hutter, ``Decoupled weight decay regularization,'' \emph{arXiv preprint arXiv:1711.05101}, 2017.

\bibitem{blip2}
J.~Li, D.~Li, S.~Savarese, and S.~Hoi, ``{BLIP-2:} bootstrapping language-image pre-training with frozen image encoders and large language models,'' in \emph{ICML}, 2023.

\bibitem{wfm}
R.~Margolin, L.~Zelnik-Manor, and A.~Tal, ``How to evaluate foreground maps?'' in \emph{CVPR}, 2014, pp. 248--255.

\bibitem{Emeasure2}
D.-P. Fan, C.~Gong, Y.~Cao, B.~Ren, M.-M. Cheng, and A.~Borji, ``Enhanced-alignment measure for binary foreground map evaluation,'' in \emph{IJCAI}.\hskip 1em plus 0.5em minus 0.4em\relax AAAI Press, 2018.

\bibitem{Emeasure}
D.-P. Fan, M.-M. Cheng, Y.~Liu, T.~Li, and A.~Borji, ``Structure-measure: A new way to evaluate foreground maps,'' in \emph{ICCV}, 2017.

\bibitem{sMeasure}
M.-M. Cheng and D.-P. Fan, ``Structure-measure: A new way to evaluate foreground maps,'' \emph{IJCV}, vol. 129, no.~9, pp. 2622--2638, 2021.

\bibitem{DUTS}
L.~Wang, H.~Lu, Y.~Wang, M.~Feng, D.~Wang, B.~Yin, and X.~Ruan, ``Learning to detect salient objects with image-level supervision,'' in \emph{CVPR}, 2017.

\bibitem{SegmMaR}
Q.~Jia, S.~Yao, Y.~Liu, X.~Fan, R.~Liu, and Z.~Luo, ``Segment, magnify and reiterate: Detecting camouflaged objects the hard way,'' in \emph{CVPR}, June 2022, pp. 4713--4722.

\bibitem{CHAMELEON}
P.~Skurowski, H.~Abdulameer, J.~B{\l}aszczyk, T.~Depta, A.~Kornacki, and P.~Kozie{\l}, ``Animal camouflage analysis: Chameleon database,'' \emph{Unpublished manuscript}, vol.~2, no.~6, p.~7, 2018.

\bibitem{CAMO}
T.-N. Le, T.~V. Nguyen, Z.~Nie, M.-T. Tran, and A.~Sugimoto, ``Anabranch network for camouflaged object segmentation,'' \emph{CVIU}, vol. 184, pp. 45--56, 2019.

\bibitem{CLandMIM}
N.~Park, W.~Kim, B.~Heo, T.~Kim, and S.~Yun, ``What do self-supervised vision transformers learn?'' in \emph{ICLR}, 2023.

\bibitem{FSPNet}
Z.~Huang, H.~Dai, T.-Z. Xiang, S.~Wang, H.-X. Chen, J.~Qin, and H.~Xiong, ``Feature shrinkage pyramid for camouflaged object detection with transformers,'' in \emph{CVPR}, June 2023, pp. 5557--5566.

\bibitem{FEDER}
C.~He, K.~Li, Y.~Zhang, L.~Tang, Y.~Zhang, Z.~Guo, and X.~Li, ``Camouflaged object detection with feature decomposition and edge reconstruction,'' in \emph{CVPR}, June 2023, pp. 22\,046--22\,055.

\bibitem{PENet}
X.~Li, J.~Yang, S.~Li, J.~Lei, J.~Zhang, and D.~Chen, ``Locate, refine and restore: A progressive enhancement network for camouflaged object detection,'' in \emph{IJCAI}, 2023, pp. 1116--1124.

\bibitem{HitNet}
X.~Hu, S.~Wang, X.~Qin, H.~Dai, W.~Ren, D.~Luo, Y.~Tai, and L.~Shao, ``High-resolution iterative feedback network for camouflaged object detection,'' in \emph{AAAI}, 2023, pp. 881--889.

\bibitem{OSFormer}
J.~Pei, T.~Cheng, D.-P. Fan, H.~Tang, C.~Chen, and L.~Van~Gool, ``Osformer: One-stage camouflaged instance segmentation with transformers,'' in \emph{ECCV}.\hskip 1em plus 0.5em minus 0.4em\relax Springer, 2022.

\bibitem{VST}
N.~Liu, N.~Zhang, K.~Wan, L.~Shao, and J.~Han, ``Visual saliency transformer,'' in \emph{ICCV}, October 2021, pp. 4722--4732.

\bibitem{ZegCLIP}
Z.~Zhou, Y.~Lei, B.~Zhang, L.~Liu, and Y.~Liu, ``Zegclip: Towards adapting clip for zero-shot semantic segmentation,'' in \emph{CVPR}, June 2023, pp. 11\,175--11\,185.

\bibitem{SwiGLUFFN}
N.~Shazeer, ``Glu variants improve transformer,'' \emph{arXiv preprint arXiv:2002.05202}, 2020.

\bibitem{RoPE}
J.~Su, M.~Ahmed, Y.~Lu, S.~Pan, W.~Bo, and Y.~Liu, ``Roformer: Enhanced transformer with rotary position embedding,'' \emph{Neurocomputing}, vol. 568, p. 127063, 2024.

\bibitem{SPARC}
I.~Bica, A.~Ili{\'c}, M.~Bauer, G.~Erdogan, M.~Bo{\v{s}}njak, C.~Kaplanis, A.~A. Gritsenko, M.~Minderer, C.~Blundell, R.~Pascanu \emph{et~al.}, ``Improving fine-grained understanding in image-text pre-training,'' \emph{arXiv preprint arXiv:2401.09865}, 2024.

\bibitem{vit}
A.~Dosovitskiy, L.~Beyer, A.~Kolesnikov, D.~Weissenborn, X.~Zhai, T.~Unterthiner, M.~Dehghani, M.~Minderer, G.~Heigold, S.~Gelly, J.~Uszkoreit, and N.~Houlsby, ``An image is worth 16x16 words: Transformers for image recognition at scale,'' \emph{ICLR}, 2021.

\bibitem{pvtv2}
W.~Wang, E.~Xie, X.~Li, D.-P. Fan, K.~Song, D.~Liang, T.~Lu, P.~Luo, and L.~Shao, ``Pvt v2: Improved baselines with pyramid vision transformer,'' \emph{Computational Visual Media}, vol.~8, no.~3, pp. 415--424, 2022.

\bibitem{pvt}
------, ``Pyramid vision transformer: A versatile backbone for dense prediction without convolutions,'' in \emph{ICCV}, 2021, pp. 568--578.

\bibitem{SegFormer}
E.~Xie, W.~Wang, Z.~Yu, A.~Anandkumar, J.~M. Alvarez, and P.~Luo, ``Segformer: Simple and efficient design for semantic segmentation with transformers,'' in \emph{NeurIPS}, vol.~34, 2021, pp. 12\,077--12\,090.

\bibitem{EVA02}
Y.~Fang, Q.~Sun, X.~Wang, T.~Huang, X.~Wang, and Y.~Cao, ``Eva-02: A visual representation for neon genesis,'' \emph{arXiv preprint arXiv:2303.11331}, 2023.

\bibitem{EVA01}
Y.~Fang, W.~Wang, B.~Xie, Q.~Sun, L.~Wu, X.~Wang, T.~Huang, X.~Wang, and Y.~Cao, ``Eva: Exploring the limits of masked visual representation learning at scale,'' in \emph{CVPR}, June 2023, pp. 19\,358--19\,369.

\bibitem{ViTDet}
Y.~Li, H.~Mao, R.~Girshick, and K.~He, ``Exploring plain vision transformer backbones for object detection,'' in \emph{Computer Vision -- ECCV 2022}, S.~Avidan, G.~Brostow, M.~Ciss{\'e}, G.~M. Farinella, and T.~Hassner, Eds.\hskip 1em plus 0.5em minus 0.4em\relax Cham: Springer Nature Switzerland, 2022, pp. 280--296.

\bibitem{bitfit}
E.~B. Zaken, S.~Ravfogel, and Y.~Goldberg, ``Bitfit: Simple parameter-efficient fine-tuning for transformer-based masked language-models,'' \emph{arXiv preprint arXiv:2106.10199}, 2022.

\bibitem{VPT}
M.~Jia, L.~Tang, B.-C. Chen, C.~Cardie, S.~Belongie, B.~Hariharan, and S.-N. Lim, ``Visual prompt tuning,'' in \emph{ECCV}, 2022, pp. 709--727.

\bibitem{LoRA}
E.~J. Hu, Y.~Shen, P.~Wallis, Z.~Allen-Zhu, Y.~Li, S.~Wang, L.~Wang, and W.~Chen, ``Lo{RA}: Low-rank adaptation of large language models,'' in \emph{ICLR}, 2022.

\bibitem{SAMAdapter}
T.~Chen, L.~Zhu, C.~Deng, R.~Cao, Y.~Wang, S.~Zhang, Z.~Li, L.~Sun, Y.~Zang, and P.~Mao, ``Sam-adapter: Adapting segment anything in underperformed scenes,'' in \emph{ICCVW}, 2023, pp. 3367--3375.

\bibitem{AdaptFormer}
S.~Chen, C.~GE, Z.~Tong, J.~Wang, Y.~Song, J.~Wang, and P.~Luo, ``Adaptformer: Adapting vision transformers for scalable visual recognition,'' in \emph{NeurIPS}, vol.~35, 2022, pp. 16\,664--16\,678.

\bibitem{MonaAdapter}
D.~Yin, L.~Hu, B.~Li, and Y.~Zhang, ``Adapter is all you need for tuning visual tasks,'' \emph{arXiv preprint arXiv:2311.15010}, 2023.

\bibitem{nlpadapter02}
J.~Pfeiffer, A.~Kamath, A.~R{\"u}ckl{\'e}, K.~Cho, and I.~Gurevych, ``{A}dapter{F}usion: Non-destructive task composition for transfer learning,'' in \emph{EACL}, 2021, pp. 487--503.

\bibitem{nlp_adapter}
N.~Houlsby, A.~Giurgiu, S.~Jastrzebski, B.~Morrone, Q.~De~Laroussilhe, A.~Gesmundo, M.~Attariyan, and S.~Gelly, ``Parameter-efficient transfer learning for {NLP},'' in \emph{ICML}, vol.~97.\hskip 1em plus 0.5em minus 0.4em\relax PMLR, 2019, pp. 2790--2799.

\bibitem{PEFT}
J.~He, C.~Zhou, X.~Ma, T.~Berg-Kirkpatrick, and G.~Neubig, ``Towards a unified view of parameter-efficient transfer learning,'' in \emph{ICLR}, 2022.

\bibitem{CODC01}
Y.~Sun, S.~Wang, C.~Chen, and T.-Z. Xiang, ``Boundary-guided camouflaged object detection,'' in \emph{IJCAI}, 2022, pp. 1335--1341.

\bibitem{CODD01}
Z.~WU, D.~P. Paudel, D.-P. Fan, J.~Wang, S.~Wang, C.~Demonceaux, R.~Timofte, and L.~Van~Gool, ``Source-free depth for object pop-out,'' in \emph{ICCV}, October 2023, pp. 1032--1042.

\bibitem{SODT02}
Q.~Zhang, N.~Huang, L.~Yao, D.~Zhang, C.~Shan, and J.~Han, ``Rgb-t salient object detection via fusing multi-level cnn features,'' \emph{IEEE TIP}, vol.~29, pp. 3321--3335, 2020.

\bibitem{SODT01}
Z.~Tu, Y.~Ma, Z.~Li, C.~Li, J.~Xu, and Y.~Liu, ``Rgbt salient object detection: A large-scale dataset and benchmark,'' \emph{IEEE TMM}, vol.~25, pp. 4163--4176, 2023.

\bibitem{SODD03}
N.~Liu, N.~Zhang, L.~Shao, and J.~Han, ``Learning selective mutual attention and contrast for rgb-d saliency detection,'' \emph{IEEE TPAMI}, vol.~44, no.~12, pp. 9026--9042, 2022.

\bibitem{SODD02CPFP}
J.-X. Zhao, Y.~Cao, D.-P. Fan, M.-M. Cheng, X.-Y. Li, and L.~Zhang, ``Contrast prior and fluid pyramid integration for rgbd salient object detection,'' in \emph{CVPR}, June 2019.

\bibitem{SODD01DPANet}
Z.~Chen, R.~Cong, Q.~Xu, and Q.~Huang, ``Dpanet: Depth potentiality-aware gated attention network for rgb-d salient object detection,'' \emph{IEEE TIP}, vol.~30, pp. 7012--7024, 2021.

\bibitem{OVCOS}
Y.~Pang, X.~Zhao, J.~Zuo, L.~Zhang, and H.~Lu, ``Open-vocabulary camouflaged object segmentation,'' in \emph{ECCV}, 2024.

\bibitem{SAM}
A.~Kirillov, E.~Mintun, N.~Ravi, H.~Mao, C.~Rolland, L.~Gustafson, T.~Xiao, S.~Whitehead, A.~C. Berg, W.-Y. Lo, P.~Doll{\'a}r, and R.~Girshick, ``Segment anything,'' in \emph{ICCV}, 2023.

\bibitem{CLIP}
A.~Radford, J.~W. Kim, C.~Hallacy, A.~Ramesh, G.~Goh, S.~Agarwal, G.~Sastry, A.~Askell, P.~Mishkin, J.~Clark, G.~Krueger, and I.~Sutskever, ``Learning transferable visual models from natural language supervision,'' in \emph{ICML}, vol. 139, 2021, pp. 8748--8763.

\bibitem{openclip}
M.~Cherti, R.~Beaumont, R.~Wightman, M.~Wortsman, G.~Ilharco, C.~Gordon, C.~Schuhmann, L.~Schmidt, and J.~Jitsev, ``Reproducible scaling laws for contrastive language-image learning,'' in \emph{CVPR}, June 2023, pp. 2818--2829.

\bibitem{ZSL}
C.~H. Lampert, H.~Nickisch, and S.~Harmeling, ``Learning to detect unseen object classes by between-class attribute transfer,'' in \emph{CVPR}, 2009, pp. 951--958.

\bibitem{Park_2017_CVPR}
J.~Park, Y.-W. Tai, D.~Cho, and I.~So~Kweon, ``A unified approach of multi-scale deep and hand-crafted features for defocus estimation,'' in \emph{CVPR}, July 2017.

\bibitem{ZoomNet}
Y.~Pang, X.~Zhao, T.-Z. Xiang, L.~Zhang, and H.~Lu, ``Zoom in and out: A mixed-scale triplet network for camouflaged object detection,'' in \emph{CVPR}, June 2022, pp. 2160--2170.

\bibitem{U2Net}
X.~Qin, Z.~Zhang, C.~Huang, M.~Dehghan, O.~Zaiane, and M.~Jagersand, ``U2-net: Going deeper with nested u-structure for salient object detection,'' \emph{PR}, vol. 106, p. 107404, 2020.

\bibitem{ZHANG201735}
Q.~Zhang, J.~Lin, Y.~Tao, W.~Li, and Y.~Shi, ``Salient object detection via color and texture cues,'' \emph{Neurocomputing}, vol. 243, pp. 35--48, 2017.

\bibitem{NC4K}
Y.~Lyu, J.~Zhang, Y.~Dai, A.~Li, B.~Liu, N.~Barnes, and D.-P. Fan, ``Simultaneously localize, segment and rank the camouflaged objects,'' in \emph{CVPR}, 2021.

\bibitem{SAM-MFG}
C.~He, K.~Li, Y.~Zhang, G.~Xu, L.~Tang, Y.~Zhang, Z.~Guo, and X.~Li, ``Weakly-supervised concealed object segmentation with sam-based pseudo labeling and multi-scale feature grouping,'' \emph{NeurIPS}, vol.~36, 2024.

\bibitem{CRNet}
R.~He, Q.~Dong, J.~Lin, and R.~W.H.~Lau, ``Weakly-supervised camouflaged object detection with scribble annotations,'' \emph{AAAI}, vol.~37, no.~1, pp. 781--789, Jun. 2023.

\bibitem{SINetV2}
D.-P. Fan, G.-P. Ji, M.-M. Cheng, and L.~Shao, ``Concealed object detection,'' \emph{IEEE TPAMI}, vol.~44, no.~10, pp. 6024--6042, 2022.

\bibitem{SINet}
D.-P. Fan, G.-P. Ji, G.~Sun, M.-M. Cheng, J.~Shen, and L.~Shao, ``Camouflaged object detection,'' in \emph{CVPR}, 2020.

\bibitem{zhu18b_eccv}
L.~Zhu, Z.~Deng, X.~Hu, C.-W. Fu, X.~Xu, J.~Qin, and P.-A. Heng, ``Bidirectional feature pyramid network with recurrent attention residual modules for shadow detection,'' in \emph{ECCV}, 2018.

\bibitem{Shi01cvpr}
J.~Shi, L.~Xu, and J.~Jia, ``Discriminative blur detection features,'' in \emph{CVPR}, 2014, pp. 2965--2972.

\bibitem{cun2020defocus}
X.~Cun and C.-M. Pun, ``Defocus blur detection via depth distillation,'' \emph{arXiv preprint arXiv:2007.08113}, 2020.

\bibitem{DeFusionNET}
C.~Tang, X.~Liu, X.~Zheng, W.~Li, J.~Xiong, L.~Wang, A.~Y. Zomaya, and A.~Longo, ``Defusionnet: Defocus blur detection via recurrently fusing and refining discriminative multi-scale deep features,'' \emph{IEEE TPAMI}, vol.~44, no.~2, pp. 955--968, 2022.

\bibitem{Wu01ACMMM}
Y.~Wu, W.~Abd-Almageed, and P.~Natarajan, ``Deep matching and validation network: An end-to-end solution to constrained image splicing localization and detection,'' in \emph{ACM MM}, 2017, p. 1480–1502.

\bibitem{Zhou_2018_CVPR}
P.~Zhou, X.~Han, V.~I. Morariu, and L.~S. Davis, ``Learning rich features for image manipulation detection,'' in \emph{CVPR}, June 2018.

\bibitem{CVPR2020_LDF}
J.~Wei, S.~Wang, Z.~Wu, C.~Su, Q.~Huang, and Q.~Tian, ``Label decoupling framework for salient object detection,'' in \emph{CVPR}, June 2020.

\bibitem{Liu21PamiPoolNet}
J.-J. Liu, Q.~Hou, Z.-A. Liu, and M.-M. Cheng, ``Poolnet+: Exploring the potential of pooling for salient object detection,'' \emph{IEEE TPAMI}, pp.~--, 2021.

\bibitem{Liu19PoolNet}
J.-J. Liu, Q.~Hou, M.-M. Cheng, J.~Feng, and J.~Jiang, ``A simple pooling-based design for real-time salient object detection,'' in \emph{CVPR}, 2019.

\bibitem{EVPV2}
{W.Liu, X.Shen, C.-M.Pun, and X.Cun}, ``Explicit visual prompting for universal foreground segmentations,'' \emph{arXiv preprint arXiv:2305.18476}, 2023.

\bibitem{ZSCOD}
H.~Li, C.-M. Feng, Y.~Xu, T.~Zhou, L.~Yao, and X.~Chang, ``Zero-shot camouflaged object detection,'' \emph{IEEE TIP}, vol.~32, pp. 5126--5137, 2023.

\bibitem{EVP}
W.~Liu, X.~Shen, C.-M. Pun, and X.~Cun, ``Explicit visual prompting for low-level structure segmentations,'' in \emph{CVPR}, June 2023, pp. 19\,434--19\,445.

\bibitem{fan2023advances}
D.-P. Fan, G.-P. Ji, P.~Xu, M.-M. Cheng, C.~Sakaridis, and L.~Van~Gool, ``Advances in deep concealed scene understanding,'' \emph{Visual Intelligence}, vol.~1, no.~1, p.~16, 2023.

\bibitem{ji2023sam}
G.-P. Ji, D.-P. Fan, P.~Xu, M.-M. Cheng, B.~Zhou, and L.~Van~Gool, ``Sam struggles in concealed scenes--empirical study on" segment anything",'' \emph{Science China Information Sciences}, vol.~66, no. 226101, 2023.

\bibitem{sun2022bgnet}
Y.~Sun, S.~Wang, C.~Chen, and T.-Z. Xiang, ``Boundary-guided camouflaged object detection,'' in \emph{IJCAI}, 2022, pp. 1335--1341.

\bibitem{wu2023popnet}
Z.~Wu, D.~P. Paudel, D.-P. Fan, J.~Wang, S.~Wang, C.~Demonceaux, R.~Timofte, and L.~Van~Gool, ``Source-free depth for object pop-out,'' in \emph{ICCV}, 2023.

\bibitem{chen2023diffusion}
Z.~Chen, R.~Gao, T.-Z. Xiang, and F.~Lin, ``Diffusion model for camouflaged object detection,'' in \emph{ECAI}, 2023.

\bibitem{zoomnext}
Y.~Pang, X.~Zhao, T.-Z. Xiang, L.~Zhang, and H.~Lu, ``Zoomnext: A unified collaborative pyramid network for camouflaged object detection,'' \emph{IEEE Transactions on Pattern Analysis and Machine Intelligence}, 2024.

\end{thebibliography}


\end{document}